\documentclass[journal]{IEEEtran} 
\usepackage[utf8]{inputenc}
\usepackage{blindtext}
\usepackage[english]{babel}
\usepackage{amsmath}
\usepackage{amssymb}
\usepackage{graphicx}
\usepackage{subcaption}
\usepackage{lscape}
\usepackage{hyperref}
\usepackage{url}
\usepackage{cite}
\usepackage{multirow}
\usepackage[dvipsnames]{xcolor}
\usepackage{tabulary}
\usepackage{pifont} 

\newcommand{\norm}[1]{\left\lVert#1\right\rVert}

\title{ORB-SLAM3: An Accurate Open-Source Library for Visual, Visual-Inertial and Multi-Map SLAM}
\author{Carlos Campos\IEEEauthorrefmark{1}, Richard Elvira\IEEEauthorrefmark{1}, Juan J. Gómez Rodríguez, José M.M. Montiel and Juan D. Tardós
\thanks{\IEEEauthorrefmark{1} Both authors contributed equally to this work.}
\thanks{The authors are with the Instituto de Investigaci\'on en Ingenier\'ia de Arag\'on (I3A), Universidad de Zaragoza, 
Mar\'ia de Luna 1, 50018 Zaragoza, Spain. E-mail: \{campos, richard, jjgomez,  josemari, tardos\}@unizar.es.} 
\thanks{This work was supported in part by the Spanish government under grants PGC2018-096367-B-I00 and DPI2017-91104-EXP, and by Aragón government under grant DGA\_T45-17R.}}

\begin{document}

\thispagestyle{empty}
\newpage
\onecolumn

\begin{center}
This paper has been accepted for publication in \textit{IEEE Transactions and Robotics}.\\
\vspace{0.75cm}
DOI: \href{https://doi.org/10.1109/TRO.2021.3075644}{\textcolor{blue}{10.1109/TRO.2021.3075644}}\\
\vspace{1.25cm}
\end{center}

©2021 IEEE.  Personal use of this material is permitted.  Permission from IEEE must be obtained for all other uses, in any current or future media, including reprinting/republishing this material for advertising or promotional purposes, creating new collective works, for resale or redistribution to servers or lists, or reuse of any copyrighted component of this work in other works.

\twocolumn

\maketitle

\begin{abstract}
This paper presents ORB-SLAM3, the first system able to perform visual, visual-inertial and multi-map SLAM with monocular, stereo and RGB-D cameras, using pin-hole and fisheye lens models.

The first main novelty is a feature-based tightly-integrated visual-inertial SLAM system that fully relies on Maximum-a-Posteriori (MAP) estimation, even during the IMU initialization phase. The result is a system that operates robustly in real time, in small and large, indoor and outdoor environments, and is two to ten times more accurate than previous approaches.

The second main novelty is a multiple map system that relies on a new place recognition method with improved recall. Thanks to it, ORB-SLAM3 is able to survive to long periods of poor visual information: when it gets lost, it starts a new map that will be seamlessly merged with previous maps when revisiting mapped areas. Compared with visual odometry systems that only use information from the last few seconds, ORB-SLAM3 is the first system able to reuse in all the algorithm stages all previous information. This allows to  include in bundle adjustment co-visible keyframes, that provide high parallax observations boosting accuracy, even if they are widely separated in time or if they come from a previous mapping session.

Our experiments show that, in all sensor configurations, ORB-SLAM3 is as robust as the best systems available in the literature, and significantly more accurate. Notably, our stereo-inertial SLAM achieves an average accuracy of 3.5\,cm in the EuRoC drone and 9\,mm under quick hand-held motions in the room of TUM-VI dataset, a setting representative of AR/VR scenarios. For the benefit of the community we make public the source code.
\end{abstract}

\section{Introduction}

Intense research on Visual Simultaneous Localization and Mapping systems (SLAM) and Visual Odometry (VO), using cameras either alone or in combination with inertial sensors, has produced during the last two decades excellent systems, with increasing accuracy and robustness. 
Modern systems rely on Maximum a Posteriori (MAP) estimation, which in the case of visual sensors corresponds to Bundle Adjustment (BA), either geometric BA that minimizes feature reprojection error, in feature-based methods, or photometric BA that minimizes the photometric error of a set of selected pixels, in direct methods. 

With the recent emergence of VO systems that integrate loop closing techniques, the frontier between VO and SLAM is more diffuse. 
The goal of Visual SLAM is to use the sensors on-board a mobile agent to build a map of the environment and compute in real-time the pose of the agent in that map. In contrast, VO systems put their focus on computing the agent's ego-motion, not on building a map. The big advantage of a SLAM map is that it allows matching and using in BA previous observations performing three types of data association (extending the terminology used in  \cite{cadena2016past}):
\begin{itemize}
    \item \textbf{Short-term data association}, matching map elements obtained during the last few seconds. This is the only data association type used by most VO systems, that forget environment elements once they get out of view, resulting in continuous estimation drift even when the system moves in the same area.
    \item \textbf{Mid-term data association}, matching map elements that are close to the camera whose accumulated drift is still small. These can be matched and used in BA in the same way than short-term observations and allow to reach zero drift when the systems moves in mapped areas. They are the key of the better accuracy obtained by our system compared against VO systems with loop detection.
    \item \textbf{Long-term data association}, matching observations with elements in previously visited areas using a place recognition technique, regardless of the accumulated drift (loop detection), the current area being previously mapped in a disconnected map (map merging), or the tracking being lost (relocalization). Long-term matching allows to reset the drift and to correct the map using pose-graph (PG) optimization, or more accurately, using BA. This is the key of SLAM accuracy in medium and large loopy environments.
\end{itemize}

In this work we build on ORB-SLAM \cite{mur2015orb, mur2017orb} and ORB-SLAM Visual-Inertial \cite{mur2017visual}, the first visual and visual-inertial systems able to take full profit of short-term, mid-term and long-term data association, reaching zero drift in mapped areas. Here we go one step further providing \textbf{multi-map data association}, which allows us to match and use in BA map elements coming from previous mapping sessions, achieving the true goal of a SLAM system: building a map that can be used later to provide accurate localization.

This is essentially a system paper, whose most important contribution is the ORB-SLAM3 library itself \cite{ORBSLAM3_source}, the most complete and accurate visual, visual-inertial and multi-map SLAM system to date (see table \ref{tab:related_work}). The main novelties of ORB-SLAM3 are:

\begin{itemize}
\item \textbf{A monocular and stereo visual-inertial SLAM system} that fully relies on Maximum-a-Posteriori (MAP) estimation, even during the IMU (Inertial Measurement Unit) initialization phase. 
The initialization method proposed was previously presented in \cite{campos2020ICRA}. Here we add its integration with ORB-SLAM visual-inertial \cite{mur2017visual}, the extension to stereo-inertial SLAM, and a thorough evaluation in public datasets. Our results show that the monocular and stereo  visual-inertial systems are extremely robust and  significantly more accurate than other visual-inertial approaches, even in sequences without loops.

\item \textbf{Improved-recall place recognition}. Many recent visual SLAM and VO systems \cite{mur2015orb, qin2018vins, Rosinol20icra-Kimera} solve place recognition using the DBoW2 bag of words library \cite{GalvezTRO12}. DBoW2 requires {\em temporal consistency}, matching three consecutive keyframes to the same area, before checking {\em geometric consistency}, boosting precision at the expense of recall. As a result, the system is too slow at closing loops and reusing previously mapped areas. We propose a novel place recognition algorithm, in which candidate keyframes are first checked for geometrical consistency, and then for {\em local consistency} with three covisible keyframes, that in most occasions are already in the map. This strategy increases recall and densifies data association improving map accuracy, at the expense of a slightly higher computational cost.

\item \textbf{ORB-SLAM Atlas}, the first complete multi-map SLAM system able to handle visual and visual-inertial systems, in monocular and stereo configurations. The Atlas can represent a set of disconnected maps, and apply to them all the mapping operations smoothly: place recognition, camera relocalization, loop closure and accurate seamless map merging.  This allows to automatically use and combine maps built at different times, performing  incremental multi-session SLAM.
A preliminary version of ORB-SLAM Atlas for visual sensors was presented in \cite{elvira2019iros}. Here we add the new place recognition system, the visual-inertial multi-map system and its evaluation on public datasets. 

\item \textbf{An abstract camera representation} making the SLAM code agnostic of the camera model used, and allowing to add new models by providing their projection, unprojection and Jacobian functions. We provide the implementations of pin-hole  \cite{tsai1987versatile} and fisheye \cite{kannala2006generic} models.

\end{itemize}

All these novelties, together with a few code improvements make ORB-SLAM3 the new reference visual and visual-inertial open-source SLAM library, being as robust as the best systems available in the literature, and significantly more accurate, as shown by our experimental results in section \ref{sec:experiments}. We also provide comparisons between monocular, stereo, monocular-inertial and stereo-inertial SLAM results that can be of interest for practitioners.

\newcommand{\doble}[2]{\begin{tabular}{@{}c@{}}#1\\#2\end{tabular}}
\newcommand{\doblel}[2]{\begin{tabular}{@{}l@{}}#1\\#2\end{tabular}}
\newcommand{\VGood}{\doble{Very}{Good}}
\newcommand{\LocalBA}{\doble{Local}{BA}}

\begin{table*}
\centering
\caption{Summary of the most representative visual (top) and visual-inertial (bottom) systems, in chronological order.}
\label{tab:related_work}
\begin{tabulary}{\textwidth}{|l|c|c|c|c|c|c|c|c|c|c|c|c|c|c|c|c|}
\hline
 & \rotatebox{90}{\textbf{\doblel{SLAM}{or VO}}} & \rotatebox{90}{\textbf{\begin{tabular}{@{}l@{}}Pixels\\used\end{tabular}}} & \rotatebox{90}{\textbf{\begin{tabular}{@{}l@{}}Data\\association\end{tabular}}} & \rotatebox{90}{\textbf{Estimation}} & \rotatebox{90}{\textbf{\doblel{Relocali-}{zation}}} & \rotatebox{90}{\textbf{\begin{tabular}{@{}l@{}}Loop\\closing\end{tabular}}} & \rotatebox{90}{\textbf{Multi Maps }} & \rotatebox{90}{\textbf{Mono}} & \rotatebox{90}{\textbf{Stereo}} & \rotatebox{90}{\textbf{Mono IMU}} & \rotatebox{90}{\textbf{Stereo IMU}} & \rotatebox{90}{\textbf{Fisheye}} & \rotatebox{90}{\textbf{Accuracy}} & \rotatebox{90}{\textbf{Robustness}} &
 \rotatebox{90}{\textbf{Open source}}\\ 
 \hline
\doblel{Mono-SLAM}{\cite{monoslam2003,monoslam}}   
& SLAM  & \doble{Shi}{Tomasi}       & Correlation          & EKF            & -             & -  
& -                  & \checkmark           & -               & -           & -             & - 
& Fair            & Fair        & \cite{monoSLAM_source2}$^1$\\ 
\hline
\doblel{PTAM}{\cite{ptam,PTAM2,ptam-phone}}         
& SLAM  & FAST          & \doble{Pyramid}{SSD}              & BA            & Thumbnail     & -         
& -                  & \checkmark            & -            & -             & -             & -
& \VGood       & Fair       & \cite{PTAM_source}      \\ 
\hline
\doblel{LSD-SLAM}{\cite{LSDSLAM,StereoLSD}}     
& SLAM  & Edgelets           & Direct               & PG            & -         & \doble{FABMAP}{PG}    
& -                  & \checkmark            & \checkmark      & -              & -         & -
& Good            &  Fair      &  \cite{LSDSLAM_source}          \\ 
\hline
SVO \cite{svo, forster2017svo}         
& VO    & \doble{FAST+}{Hi.grad.}                 & Direct               & \LocalBA            & -             & - 
& -         & \checkmark        & \checkmark            & -                 & -             & \checkmark 
& \VGood        & \VGood          &  \cite{SVO_source}$^2$      \\ 
\hline
\doblel{ORB-SLAM2}{\cite{mur2015orb,mur2017orb}}
& SLAM  & ORB                  & Descriptor           & \LocalBA        & DBoW2     & \doble{DBoW2}{PG+BA}
& -         & \checkmark        & \checkmark            & -             & -                 & -
& Exc.          & \VGood         &  \cite{ORBSLAM2_source}        \\ 
\hline
DSO \cite{Engel-et-al-pami2018, matsuki2018omnidirectional, wang2017stereoDSO}
& VO    & \doble{High}{grad.}          & Direct         & \LocalBA      & -             & -  
& -         & \checkmark            & \checkmark       & -               & -             & \checkmark 
& Fair       & \VGood    &  \cite{DSO_source}     \\ 
\hline
DSM \cite{zubizarreta2020direct}         
& SLAM  & \doble{High}{grad.}          & Direct         & \LocalBA      & -             & -  
& -         & \checkmark            & -               & -               & -             & -
& \VGood       & \VGood    & \cite{DSM_source}     \\ 
\hline 
\hline
\doblel{MSCKF}{\cite{mourikis2007multi,li2013high,paul2017comparative, paul2018alternating}}
& VO    & \doble{Shi}{Tomasi}   & \doble{Cross}{correlation}    & EKF           & -             & - 
& -             & \checkmark                & -               & \checkmark      & \checkmark    & -     
& Fair      & \VGood     & \cite{MSCKF_source}$^3$    \\ 
\hline
\doblel{OKVIS}{\cite{leutenegger2013keyframe,leutenegger2015keyframe}}
& VO    & BRISK             & Descriptor         & \LocalBA            & -          & -         & - 
& -             & -               & \checkmark                & \checkmark          &  \checkmark                 
& Good            & \VGood  &  \cite{OKVIS_source}       \\ 
\hline
\doblel{ROVIO}{\cite{bloeschIROS15,bloesch2017iterated}} 
& VO    & \doble{Shi}{Tomasi}           & Direct               & EKF        & -             & -       
& -             & -             & -               & \checkmark              &  \checkmark              &  \checkmark 
& Good            & \VGood &   \cite{ROVIO_source} \\ 
\hline
\doblel{ORBSLAM-VI}{\cite{mur2017visual}}   
& SLAM  & ORB           & Descriptor           & \LocalBA       & DBoW2     & \doble{DBoW2}{PG+BA}  
& -         & \checkmark    & -       & \checkmark    & -         & -
& \VGood       & \VGood  & -      \\ 
\hline
\doblel{VINS-Fusion}{\cite{qin2018vins, qin2019general}}
& VO    & \doble{Shi}{Tomasi}       & KLT       & \LocalBA      & DBoW2     & \doble{DBoW2}{PG}
& \checkmark        & -             & \checkmark        & \checkmark        & \checkmark    & \checkmark
& Good       & Exc.     &  \cite{VINS-Fusion_source}       \\ 
\hline
VI-DSO \cite{stumberg2018direct}      
& VO    & \doble{High}{grad.}        & Direct         & \LocalBA        & -             & -                
& -             & -             & -               & \checkmark         & -              & -                
& \VGood       & Exc.      & -        \\ 
\hline
\doblel{BASALT}{\cite{usenko2020visual}} 
& VO  & FAST      & \doble{KLT}{(LSSD)}     & \LocalBA        & -              & \doble{ORB}{BA}
& -             & -             & -               & -           & \checkmark    & \checkmark    
& \VGood        & Exc.     & \cite{BASALT_source}   \\ 
\hline
Kimera \cite{Rosinol20icra-Kimera}      
& VO  & \doble{Shi}{Tomasi}       & KLT           & \LocalBA      & -           & \doble{DBoW2}{PG}
& -             & -             & -               & -           & \checkmark    & -      
& Good          & Exc.      & \cite{Kimera_source}   \\ 
\hline
\doblel{ORB-SLAM3}{(ours)}
& SLAM  & ORB           & Descriptor         & \LocalBA     & DBoW2             & \doble{DBoW2}{PG+BA}  
& \checkmark        & \checkmark    & \checkmark        & \checkmark    & \checkmark        & \checkmark
& Exc.            & Exc.     &  \cite{ORBSLAM3_source} \\ \hline
\multicolumn{16}{l}{\begin{tabular}[l]{@{}l@{}}
\\
$^1$ Last source code provided by a different author. Original software is available at \cite{monoSLAM_source}. \\ 
$^2$ Source code available only for the first version, SVO 2.0 is not open source. \\
$^3$ MSCKF is patented \cite{roumeliotis2017patent}, only a re-implementation by a different author is available as open source. \end{tabular} }
\end{tabulary}
\end{table*}

\section{Related Work}
Table \ref{tab:related_work} presents a summary of the most representative visual and visual-inertial systems, showing the main techniques used for estimation and data association. The qualitative accuracy and robustness ratings included in the table are based on the results presented in section \ref{sec:experiments}, and the comparison between PTAM, LSD-SLAM and ORB-SLAM reported in \cite{mur2015orb}.

\subsection{Visual SLAM}

Monocular SLAM was first solved in MonoSLAM \cite{monoslam2003, monoslam, Civera2008} using an Extended Kalman Filter (EKF) and Shi-Tomasi points that were tracked in subsequent images doing a guided search by correlation. Mid-term data association was significantly improved using techniques that guarantee that the feature matches used are consistent, achieving hand-held visual SLAM \cite{Clemente-RSS07,civera2010}.

In contrast, keyframe-based approaches estimate the map using only a few selected frames, discarding the information coming from intermediate frames. This allows to perform the more costly, but more accurate, BA optimization at keyframe rate. The most representative system was PTAM \cite{ptam}, that split camera tracking and mapping in two parallel threads. Keyframe-based techniques are more accurate than filtering for the same computational cost  \cite{whyFilter}, becoming the gold standard in visual SLAM and VO. Large scale monocular SLAM was achieved in \cite{strasdat2010scale} using sliding-window BA, and in \cite{DWO} using a double-window optimization and a covisibility graph. 

Building on these ideas, ORB-SLAM \cite{mur2015orb,mur2017orb} uses ORB features, whose descriptor provides short-term and mid-term data association, builds a covisibility graph to limit the complexity of tracking and mapping, and performs loop closing and relocalization using the bag-of-words library DBoW2 \cite{GalvezTRO12}, achieving long-term data association. To date is the only visual SLAM system integrating the three types of data association, which we believe is the key of its excellent accuracy. In this work we improve its robustness in pure visual SLAM with the new Atlas system that starts a new map when tracking is lost, and its accuracy in loopy scenarios with the new place recognition method with improved recall. 

Direct methods do not extract features, but use directly the pixel intensities in the images, and estimate motion and structure by minimizing a photometric error. LSD-SLAM \cite{LSDSLAM} was able to build large scale semi-dense maps using high gradient pixels. However, map estimation was reduced to pose-graph (PG) optimization, achieving lower accuracy than PTAM and ORB-SLAM \cite{mur2015orb}. The hybrid system SVO \cite{svo, forster2017svo}  extracts  FAST features, uses a direct method to track features and any pixel with nonzero intensity gradient from frame to frame, and optimizes camera trajectory and 3D structure using reprojection error. SVO is extremely efficient but, being a pure VO method, it only performs short-term data association, which limits its accuracy. Direct Sparse Odometry DSO \cite{Engel-et-al-pami2018} is able to compute accurate camera poses in situations where point detectors perform poorly, enhancing robustness in low textured areas or against blurred images. It introduces local photometric BA that simultaneously optimizes a window of seven recent keyframes and the inverse depth of the points. Extensions of this work include stereo \cite{wang2017stereoDSO}, loop closing using features and DBoW2 \cite{gao2018ldso} \cite{lee2018loosely}, and visual-inertial odometry \cite{stumberg2018direct}. Direct Sparse Mapping DSM \cite{zubizarreta2020direct} introduces the idea of map reusing in direct methods, showing the importance of mid-term data association. In all cases, the lack of integration of short, mid, and long-term data association results in lower accuracy than our proposal (see section \ref{sec:experiments}).

\subsection{Visual-Inertial SLAM}
The combination of visual and inertial sensors provide robustness to poor texture, motion blur and occlusions, and in the case of monocular systems, make scale observable.

Research in tightly coupled approaches can be traced back to MSCKF \cite{mourikis2007multi} where the EKF quadratic cost in the number of  features is avoided by feature marginalization. The initial system was perfected in \cite{li2013high} and extended to stereo in \cite{paul2017comparative, paul2018alternating}. The first tightly coupled visual odometry system based on keyframes and bundle adjustment was OKVIS \cite{leutenegger2013keyframe, leutenegger2015keyframe} that is also able to use  monocular and stereo vision. While these systems rely on features,  ROVIO \cite{bloeschIROS15, bloesch2017iterated} feeds an EFK with photometric error using direct data association. 

ORB-SLAM-VI \cite{mur2017visual} presented for the first time a visual-inertial SLAM system able to reuse a map with short-term, mid-term and long-term data association, using them in an accurate local visual-inertial BA based on IMU preintegration \cite{lupton2012visual,forster2017manifold}. However, its IMU initialization technique was too slow, taking  15 seconds, which harmed robustness and accuracy. Faster initialization techniques were proposed in \cite{martinelli2014closed,kaiser2017simultaneous}, based on a closed-form solution to jointly retrieve scale, gravity, accelerometer bias and initial velocity, as well as visual features depth. Crucially, they ignore IMU noise properties, and minimize the 3D error of points in space, and not their reprojection errors, that is the gold standard in feature-based computer vision. Our previous work \cite{campos2019} shows that this results in large unpredictable errors.

VINS-Mono \cite{qin2018vins} is a very accurate and robust monocular-inertial odometry system, with loop closing that uses DBoW2 and 4 DoF pose-graph optimization, and map-merging. Feature tracking is performed with Lucas-Kanade tracker, being slightly more robust than descriptor matching. In VINS-Fusion \cite{qin2019general} it has been extended to stereo and stereo-inertial. 

VI-DSO \cite{stumberg2018direct} extends DSO to visual-inertial odometry, proposing a bundle adjustment that combines inertial observations with the photometric error of selected high gradient pixels, what renders very good accuracy. As the information from high gradient pixels is successfully exploited, the robustness in scene regions with poor texture is also boosted. Their initialization method relies on visual-inertial BA and takes 20-30 seconds to converge within 1\% scale error.

The recent BASALT \cite{usenko2020visual} is a stereo-inertial odometry system that extracts non-linear factors from visual-inertial odometry to use them in BA, and closes loops matching ORB features, achieving very good to excellent accuracy. Kimera \cite{Rosinol20icra-Kimera} is a novel outstanding metric-semantic mapping system, but its metric part consists in stereo-inertial odometry plus loop closing with DBoW2 and pose-graph optimization, achieving similar accuracy to VINS-Fusion. 

In this work we build on ORB-SLAM-VI and extend it to stereo-inertial SLAM. We propose a novel fast initialization method based on Maximum-a-Posteriori (MAP) estimation that properly takes into account visual and inertial sensor uncertainties, and estimates the true scale with 5\% error in 2 seconds, converging to 1\% scale error in 15 seconds.  All other systems discussed above are visual-inertial odometry methods, some of them extended with loop closing, and lack the capability of using mid-term data associations. We believe that this, together with our fast and precise initialization, is the key of the better accuracy consistently obtained by our system, even in sequences without loops. 

\subsection{Multi-Map SLAM}

The idea of adding robustness to tracking losses during exploration by means of map creation and fusion was first proposed in \cite{eadeBMVC2008} within a filtering approach. One of the first keyframe-based multi-map systems was \cite{Castle2008}, but the map initialization was manual, and the system was not able to merge or relate the different sub-maps. Multi-map capability has been researched as a component of collaborative mapping systems, with several mapping agents and a central server that only receives information \cite{forster2013collaborative} or with bidirectional information flow as in C2TAM \cite{riazuelo2014c2tam}. MOARSLAM \cite{morrison2016moarslam} proposed a robust stateless client-server architecture for collaborative multi-device SLAM, but the main focus was the software architecture and did not report accuracy results.

More recently, CCM-SLAM \cite{schmuck2017multi,schmuck2018ccm} proposes a distributed multi-map system for multiple drones with bidirectional information flow, built on top of ORB-SLAM. Their focus is on overcoming the challenges of limited bandwidth and distributed processing, while ours is on accuracy and robustness, achieving significantly better results on the EuRoC dataset. SLAMM \cite{daoud2018slamm} also proposes a multi-map extension of ORB-SLAM2, but keeps sub-maps as separated entities, while we perform seamless map merging, building a more accurate global map. 

VINS-Mono \cite{qin2018vins} is a visual odometry system with loop closing and multi-map capabilities that rely on the place recognition library DBoW2 \cite{GalvezTRO12}. Our experiments show that ORB-SLAM3 is 2.6 times more accurate than VINS-Mono in monocular-inertial single-session operation on the EuRoc dataset, thanks to the ability to use mid-term data association. Our Atlas system also builds on DBoW2, but proposes a novel higher-recall place recognition technique, and performs more detailed and accurate map merging using local BA, increasing the advantage to 3.2 times better accuracy than VINS-Mono in multi-session operation on EuRoC.


\section{System Overview}

\begin{figure}
\centering
  \includegraphics[width=\columnwidth]{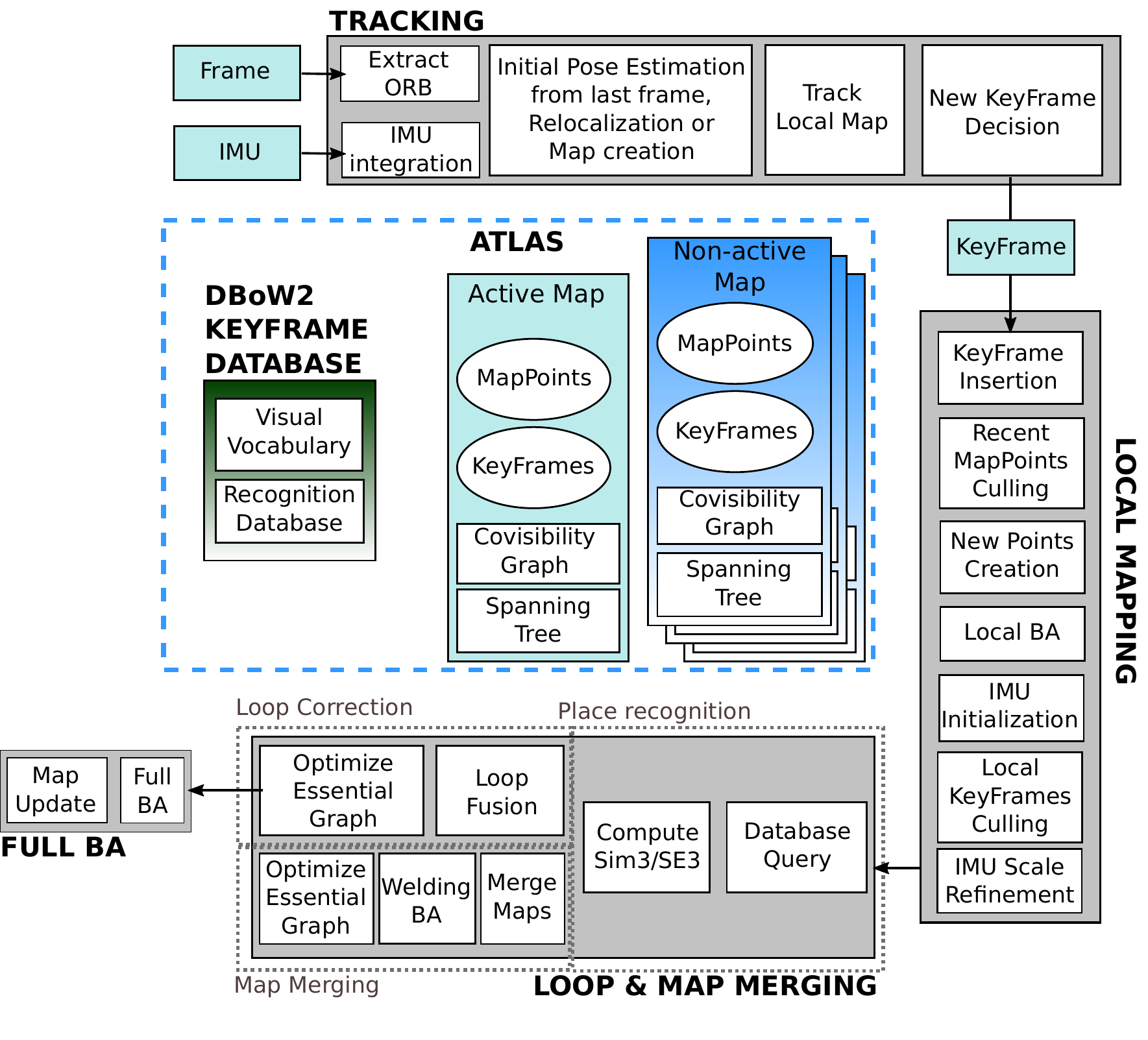}
  \caption{Main system components of ORB-SLAM3.}
  \label{fig:orbslam3_workflow}
\end{figure}

ORB-SLAM3 is built on ORB-SLAM2 \cite{mur2017orb} and ORB-SLAM-VI \cite{mur2017visual}. It is a full multi-map and multi-session system able to work in pure visual or visual-inertial modes with monocular, stereo or RGB-D sensors, using pin-hole and fisheye camera models. Figure\,\ref{fig:orbslam3_workflow} shows the main system components, that are parallel to those of ORB-SLAM2 with some significant novelties, that are summarized next:

\begin{itemize}
    \item \textbf{Atlas} is a multi-map representation composed of a set of disconnected maps. There is an active map where the tracking thread localizes the incoming frames, and is continuously optimized and grown with new keyframes by the local mapping thread. We refer to the other maps in the Atlas as the non-active maps. The system builds a unique DBoW2 database of keyframes that is used for relocalization, loop closing and map merging. 
     
    \item \textbf{Tracking thread} processes sensor information and computes the pose of the current frame with respect to the active map in real-time, minimizing the reprojection error of the matched map features. It also decides whether the current frame becomes a keyframe. In visual-inertial mode, the body velocity and IMU biases are estimated by including the inertial residuals in the optimization. When tracking is lost, the tracking thread tries to relocalize the current frame in all the Atlas' maps. If relocalized, tracking is resumed, switching the active map if needed. Otherwise, after a certain time, the active map is stored as non-active, and a new active map is initialized from scratch. 
    
    \item \textbf{Local mapping thread} adds keyframes and points to the active map, removes the redundant ones, and refines the map using visual or visual-inertial bundle adjustment, operating in a local window of keyframes close to the current frame. Additionally, in the inertial case, the IMU parameters are initialized and refined by the mapping thread using our novel MAP-estimation technique.
    
    \item \textbf{Loop and map merging thread} detects common regions between the active map and the
    whole Atlas at keyframe rate. If the common area belongs to the active map, it performs loop correction; if it belongs to a different map, both maps are seamlessly merged into a single one, that becomes the active map.
    After a loop correction, a full BA is launched in an independent thread to further refine the map without affecting real-time performance.
\end{itemize}


\section{Camera Model}
ORB-SLAM assumed in all system components a pin-hole camera model. Our goal is to abstract the camera model from the whole SLAM pipeline by extracting all properties and functions related to the camera model (projection and unprojection functions, Jacobian, etc.) into separate modules. This allows our system to use any camera model by providing the corresponding camera module. In ORB-SLAM3 library, apart from the pin-hole model, we provide the Kannala-Brandt \cite{kannala2006generic} fisheye model. 

As most popular computer vision algorithms assume a pin-hole camera model, many SLAM systems rectify either the whole image, or the feature coordinates, to work in an ideal planar retina. However, this approach is problematic for fisheye lenses, that can reach or surpass a field of view (FOV) of 180 degrees. Image rectification is not an option as objects in the periphery get enlarged and objects in the center loose resolution, hindering feature matching. Rectifying the feature coordinates requires using less than 180 degrees FOV and causes trouble to many computer vision algorithms that assume uniform reprojection error along the image, which is far from true in rectified fisheye images. This forces to crop-out the outer parts of the image, losing the advantages of large FOV: faster mapping of the environment and better robustness to occlusions. Next, we discuss how to overcome these difficulties.

\subsection{Relocalization}
A robust SLAM system needs the capability of relocalizing the camera when tracking fails. ORB-SLAM solves the relocalization problem by setting a Perspective-n-Points solver based on the ePnP algorithm \cite{lepetit2009epnp}, which assumes a calibrated pin-hole camera along all its formulation. To follow up with our approach, we need a PnP algorithm that works independently of the camera model used. For that reason, we have adopted Maximum Likelihood Perspective-n-Point algorithm (MLPnP)\cite{urban2016mlpnp} that is completely decoupled from the camera model as it uses projective rays as input. The camera model just needs to provide an unprojection function passing from pixels to projection rays, to be able to use relocalization.

\subsection{Non-rectified Stereo SLAM}
Most stereo SLAM systems assume that stereo frames are rectified, i.e. both images are transformed to pin-hole projections using the same focal length, with image planes co-planar, and are aligned with horizontal epipolar lines, such that a feature in one image can be easily matched by looking at the same row in the other image. However the assumption of rectified stereo images is very restrictive and, in many applications, is neither suitable nor feasible. For example, rectifying a divergent stereo pair, or a stereo fisheye camera would require severe image cropping, loosing the advantages of a large FOV.  

For that reason, our system does not rely on image rectification, considering the stereo rig as two monocular cameras having:

\begin{enumerate}
\item a constant relative $\textrm{SE}(3)$ transformation between them, and
\item optionally, a common image region that observes the same portion of the scene.
\end{enumerate}

These constrains allow us to effectively estimate the scale of the map by introducing that information when triangulating new landmarks and in the bundle adjustment optimization. Following up with this idea, our SLAM pipeline estimates a 6 DoF rigid body pose, whose reference system can be located in one of the cameras or in the IMU sensor, and represents the cameras with respect to the rigid body pose. 

If both cameras have an overlapping area in which we have stereo observations, we can triangulate true scale landmarks the first time they are seen. The rest of both images still has a lot of relevant information that is used as monocular information in the SLAM pipeline. Features first seen in these areas are triangulated from multiple views, as in the monocular case.

\section{Visual-Inertial SLAM}
ORB-SLAM-VI \cite{mur2017visual} was the first true visual-inertial SLAM system capable of map reusing. However, it was limited to pin-hole monocular cameras, and its initialization was too slow, failing in some challenging scenarios. In this work, we build on ORB-SLAM-VI providing a fast an accurate IMU initialization technique, and an open-source SLAM library capable of monocular-inertial and stereo-inertial SLAM, with pin-hole and fisheye cameras. 

\subsection{Fundamentals}
\label{subsec:Fundamentals}
While in pure visual SLAM, the estimated state only includes the current camera pose, in visual-inertial SLAM, additional variables need to be computed. These are the body pose $\mathbf{T}_i = [\mathbf{R}_i, \mathbf{p}_i] \in \text{SE}(3)$ and velocity $\textbf{v}_i$ in the world frame, and the gyroscope and accelerometer biases, $\textbf{b}^g_i$ and $\textbf{b}^a_i$, which are assumed to evolve according to a Brownian motion. This leads to the state vector:
\begin{equation}
    \mathcal{S}_i \doteq \{\mathbf{T}_i, \mathbf{v}_i, \mathbf{b}^g_i, \mathbf{b}^a_i\}
\end{equation}

For visual-inertial SLAM, we preintegrate IMU measurements between consecutive visual frames, $i$ and $i+1$, following the theory developed in \cite{lupton2012visual}, and formulated on manifolds in \cite{forster2017manifold}. We obtain preintegrated rotation, velocity and position measurements, denoted as $\Delta \textbf{R}_{i,i+1}$, $\Delta\textbf{v}_{i,i+1}$ and $\Delta\textbf{p}_{i,i+1}$, as well a covariance matrix $\Sigma_{\mathcal{I}_{i,i+1}}$ for the whole measurement vector. Given these preintegrated terms and states $\mathcal{S}_i$ and $\mathcal{S}_{i+1}$, we adopt the definition of inertial residual $\mathbf{r}_{\mathcal{I}_{i,i+1}}$ from \cite{forster2017manifold}:
\begin{equation}
\begin{split}
\mathbf{r}_{\mathcal{I}_{i,i+1}}& =[\textbf{r}_{\Delta \text{R}_{i,i+1}}, \textbf{r}_{\Delta \text{v}_{i,i+1}}, \textbf{r}_{\Delta \text{p}_{i,i+1}}]\\
\textbf{r}_{\Delta \text{R}_{i,i+1}}&  = \text{Log}\left(  \Delta \textbf{R}_{i,i+1}^\text{T} \textbf{R}_i^\text{T} \textbf{R}_{i+1}\right)\\
\textbf{r}_{\Delta \text{v}_{i,i+1}}&  = \textbf{R}_i^\text{T} \left( \textbf{v}_{i+1} - \textbf{v}_i - \textbf{g} \Delta t_{i,i+1}\right) - \Delta\textbf{v}_{i,i+1} \\
\textbf{r}_{\Delta \text{p}_{i,i+1}} & = \textbf{R}_i^\text{T} \left( {\textbf{p}_j} - {\textbf{p}}_i - {\textbf{v}}_i \Delta t_{i,i+1} - \frac{1}{2}\textbf{g} \Delta t^2\right) - \Delta\textbf{p}_{i,i+1}
\end{split}
\end{equation}
where $\text{Log}: \text{SO}(3) \rightarrow \mathbb{R}^3$ maps from the Lie group to the vector space.
Together with inertial residuals, we also use reprojection errors $\mathbf{r}_{ij}$ between frame $i$ and 3D point $j$ at position $\mathbf{x}_j$:
\begin{equation}
    \mathbf{r}_{ij} = \mathbf{u}_{ij}-\Pi\left(\mathbf{T}_{\text{CB}}\mathbf{T}_i^{-1} \oplus \mathbf{x}_j\right)
\end{equation}
where $\Pi: \mathbb{R}^3 \rightarrow \mathbb{R}^n$ is the projection function for the corresponding camera model, $\mathbf{u}_{ij}$ is the observation of point $j$ at image $i$, having a covariance matrix $\Sigma_{ij}$,  $\mathbf{T}_{\text{CB}} \in \text{SE}(3)$ stands for the rigid transformation from body-IMU to camera (left or right), known from calibration, and $\oplus$ is the transformation operation of $\text{SE}(3)$ group over $\mathbb{R}^3$ elements.

\begin{figure*}
\begin{subfigure}{0.40\columnwidth}
  \centering
  \includegraphics[width=1\columnwidth]{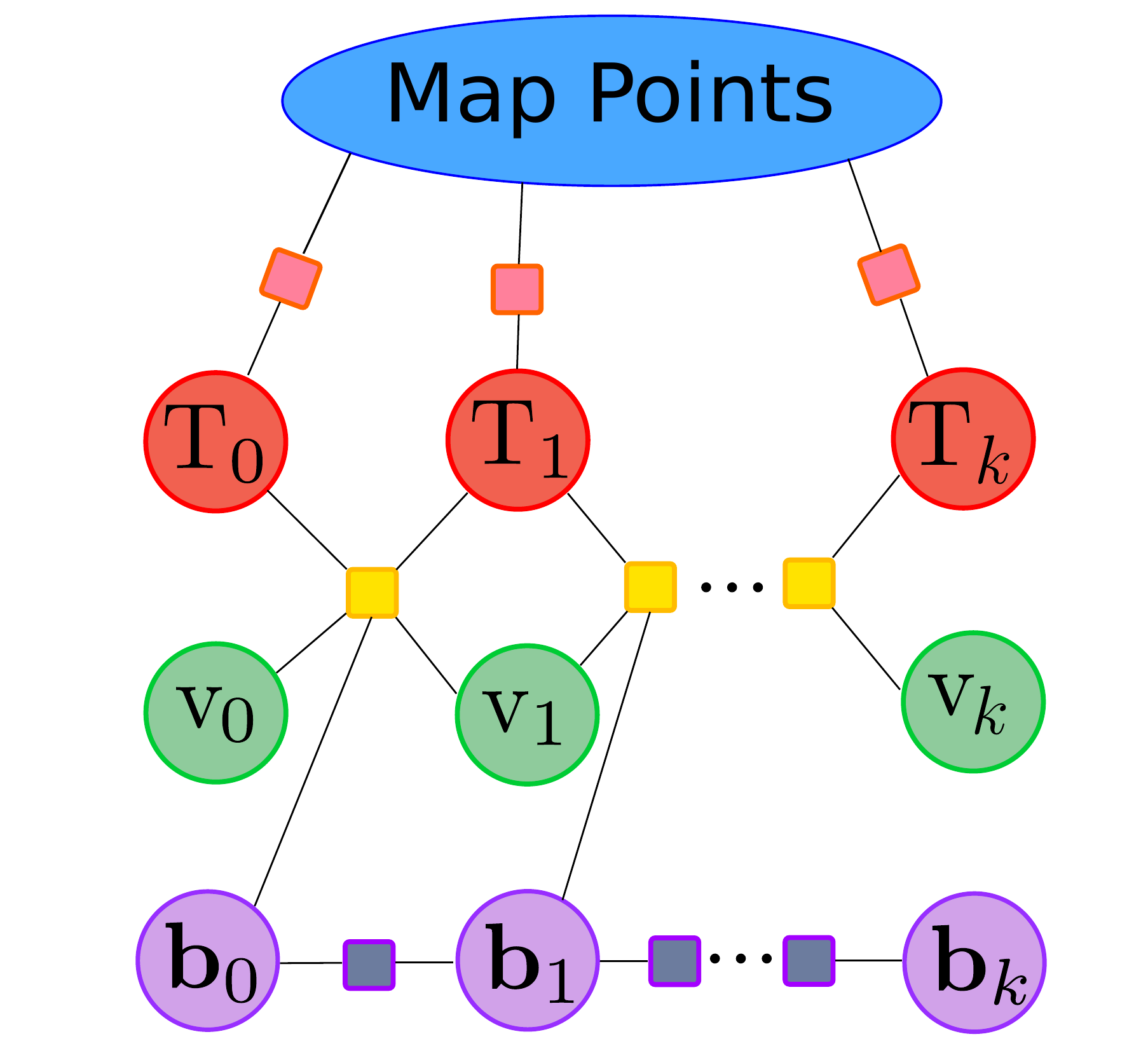}
  \caption{Visual-Inertial}
  \label{fig:VIgraph}
\end{subfigure}
\begin{subfigure}{0.40\columnwidth}
  \centering
  \includegraphics[width=1\columnwidth]{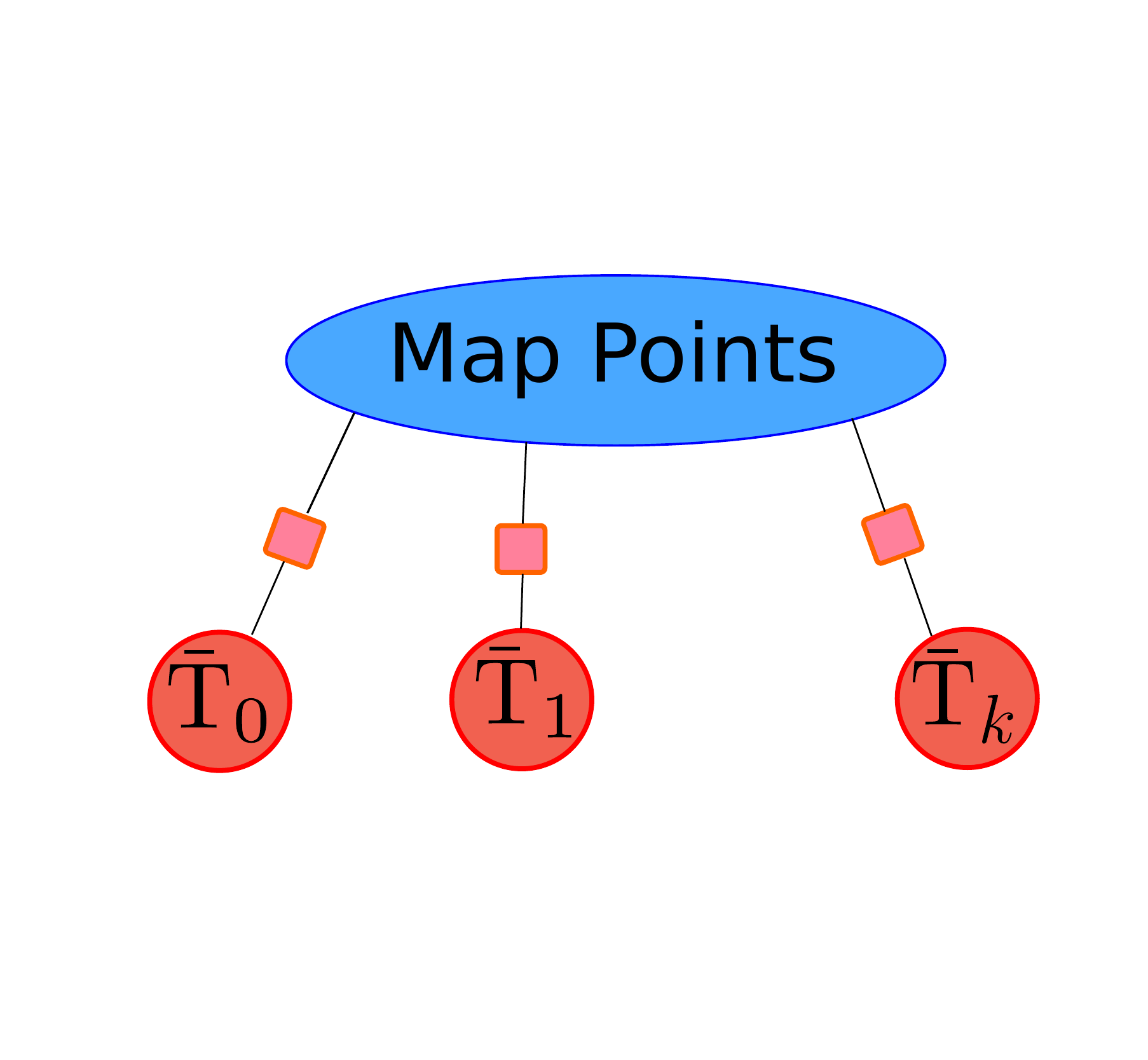}
  \caption{Visual-Only }
  \label{fig:Vgraph}
\end{subfigure}
\begin{subfigure}{0.40\columnwidth}
  \centering
  \includegraphics[width=1\columnwidth]{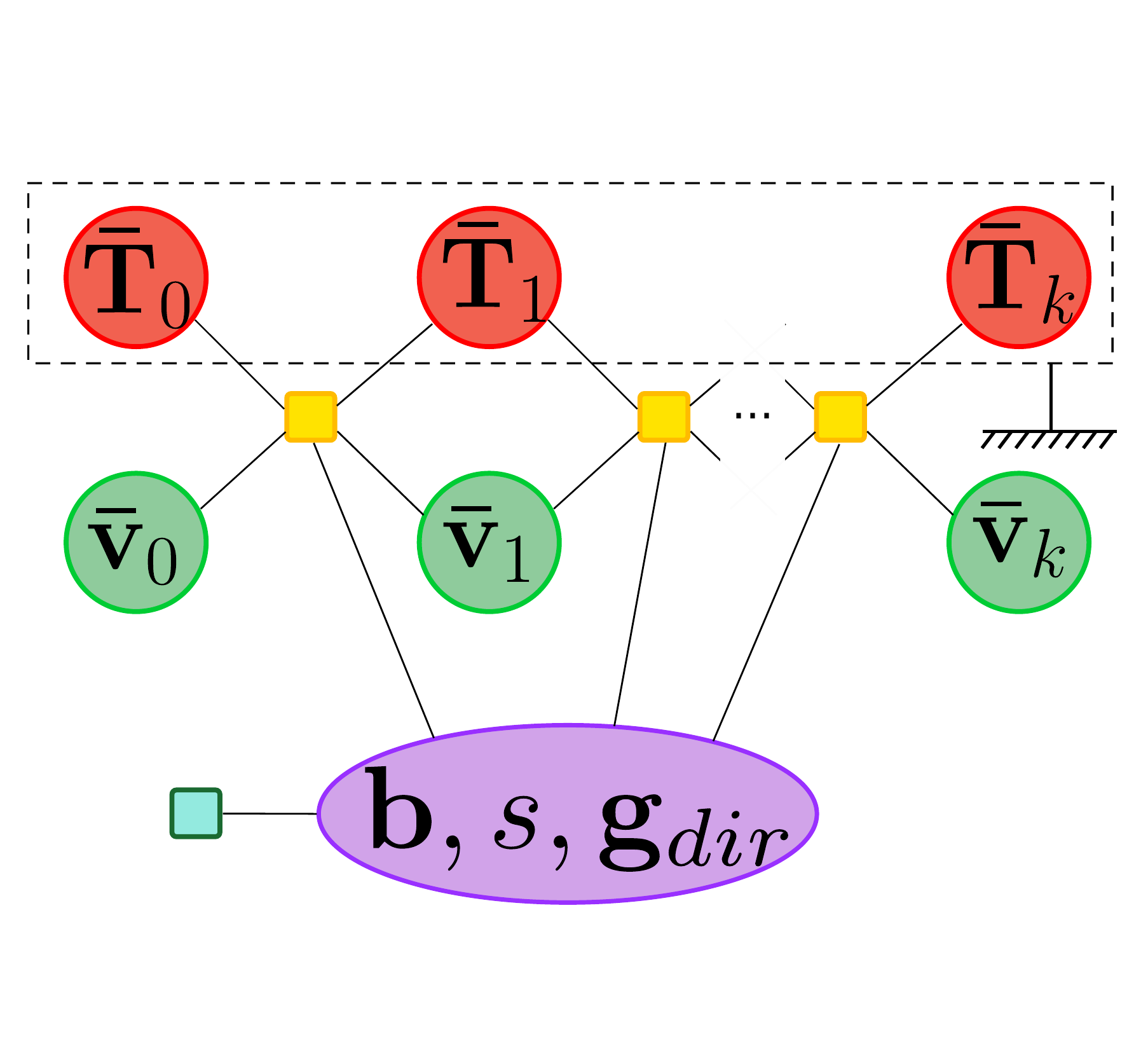}
  \caption{Inertial-Only }
  \label{fig:Igraph}
\end{subfigure}
\begin{subfigure}{0.40\columnwidth}
  \centering
  \includegraphics[width=1\columnwidth]{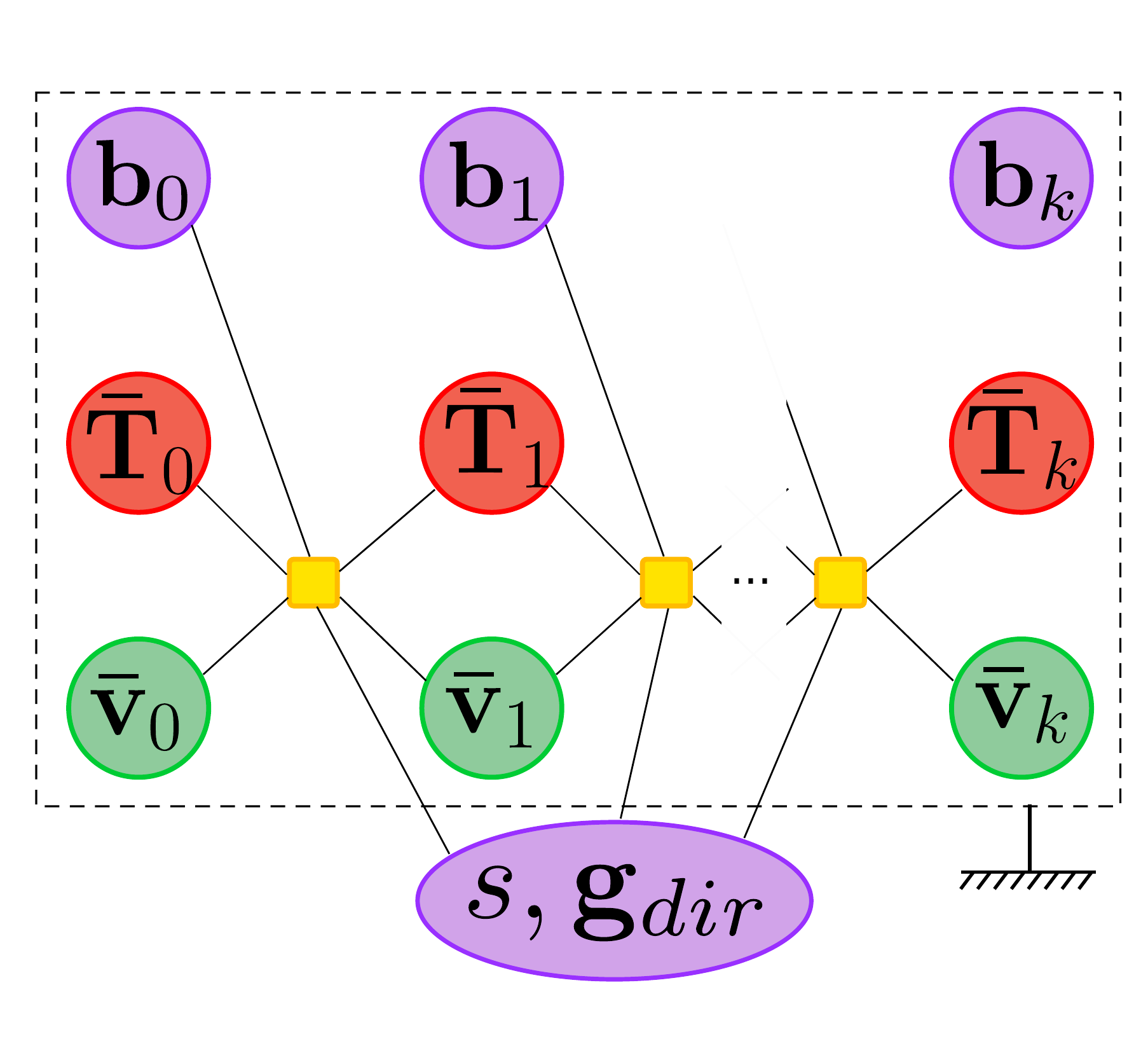}
  \caption{Scale and Gravity }
  \label{fig:SGgraph}
\end{subfigure}
\begin{subfigure}{0.40\columnwidth}
  \centering
  \includegraphics[width=1\columnwidth]{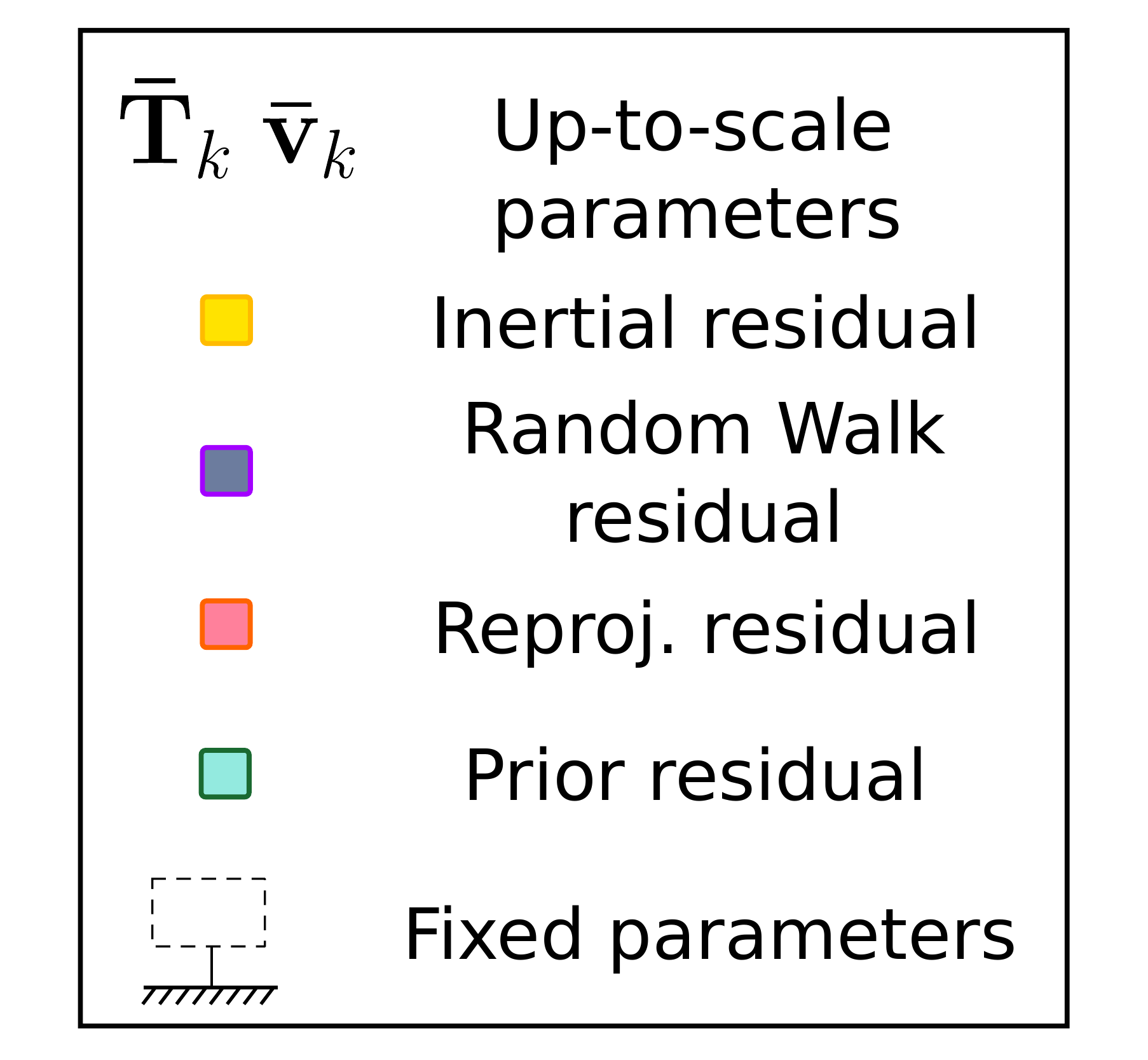}
\end{subfigure}
\caption{Factor graph representation for different optimizations along the system}
\label{fig:factor_graphs}
\end{figure*}

Combining inertial and visual residual terms, visual-inertial SLAM can be posed as a keyframe-based minimization problem  \cite{leutenegger2015keyframe}. Given a set of $k+1$ keyframes and its state $\bar{\mathcal{S}}_{k} \doteq \{\mathcal{S}_{0} \dots \mathcal{S}_{k}\}$, and  a set of $l$ 3D points and its state $\mathcal{X} \doteq \{ \mathbf{x}_0 \dots \mathbf{x}_{l-1} \}$, the visual-inertial optimization problem can be stated as:
\begin{equation}
\label{eq:imu_BA}
    \underset{\bar{\mathcal{S}}_{k} , \mathcal{X}}{\min} \left( \sum_{i = 1}^{k} \norm{\mathbf{r}_{\mathcal{I}_{i-1,i}}}_{\Sigma^{-1}_{\mathcal{I}_{i,i+1}}}^2 + \sum_{j = 0}^{l-1} \sum_{i \in \mathcal{K}^{j}} \rho_{\text{Hub}}\left(\norm{\mathbf{r}_{ij}}_{\Sigma^{-1}_{ij}}\right) \right)
\end{equation}
where $\mathcal{K}^j$ is the set of keyframes observing 3D point $j$. This optimization may be outlined as the factor-graph shown in figure \ref{fig:VIgraph}. Note that for reprojection error we use a robust Huber kernel $\rho_{\text{Hub}}$ to reduce the influence of spurious matchings, while for inertial residuals it is not needed, since miss-associations do not exist. This optimization needs to be adapted for efficiency during tracking and mapping, but more importantly, it requires good initial seeds to  converge to accurate solutions.

\subsection{IMU Initialization}
The goal of this step is to obtain good initial values for the inertial variables: body velocities, gravity direction, and IMU biases. Some systems like VI-DSO \cite{stumberg2018direct} try to solve from scratch visual-inertial BA, sidestepping a specific initialization process, obtaining slow convergence for inertial parameters (up to 30 seconds).

In this work we propose a fast and accurate initialization method based on three key insights:
\begin{itemize}
\item Pure monocular SLAM can provide very accurate initial maps \cite{mur2015orb}, whose main problem is that scale is unknown. Solving first the vision-only problem will enhance IMU initialization.
\item As shown in \cite{strasdat2010scale}, scale converges much faster when it is explicitly represented as an optimization variable, instead of using the implicit representation of BA.
\item Ignoring sensor uncertainties during IMU initialization produces large unpredictable errors \cite{campos2019}.
\end{itemize}

So, taking properly into account sensor uncertainties, we state the IMU initialization as a MAP estimation problem, split in three steps:

\begin{enumerate}
    \item {\bf Vision-only MAP Estimation}: We initialize pure monocular SLAM \cite{mur2015orb} and run it during 2 seconds, inserting keyframes at 4Hz. After this period, we have an up-to-scale map composed of $k=10$ camera poses and hundreds of points, that is optimized using visual-only BA (figure \ref{fig:Vgraph}). These poses are transformed to body reference, obtaining the trajectory  $\mathbf{\bar{T}}_{0:k}=[\mathbf{R},\mathbf{\bar{p}}]_{0:k}$ where the bar denotes up-to-scale variables in the monocular case.
    
    \item {\bf Inertial-only MAP Estimation}: In this step we aim to obtain the optimal estimation of the inertial variables, in the sense of MAP estimation, using only $\mathbf{\bar{T}}_{0:k}$ and inertial measurements between these keyframes. These inertial variables may be stacked in the inertial-only state vector:
    \begin{equation}
        \mathcal{Y}_k=\{s, \textbf{R}_{\text{w}g}, \mathbf{b}, \mathbf{\bar{v}}_{0:k} \}
    \end{equation}
    where $s\in \mathbb{R}^+$ is the scale factor of the vision-only solution; 
     $\textbf{R}_{\text{w}g} \in \text{SO}(3)$ is a rotation matrix, used to compute gravity vector $\textbf{g}$ in the world reference as $\textbf{g}=\textbf{R}_{\text{w}g}\textbf{g}_{\text{I}}$ , where $\textbf{g}_{\text{I}}=(0,0,G)^{\text{T}}$ and $G$ is the gravity magnitude;  $\mathbf{b}=(\mathbf{b}^a,\mathbf{b}^g) \in \mathbb{R}^6$ are the accelerometer and gyroscope biases assumed to be constant during initialization; and  $\mathbf{\bar{v}}_{0:k} \in \mathbb{R}^3$ is the up-to-scale body velocities from first to last keyframe, initially estimated from $\mathbf{\bar{T}}_{0:k}$. At this point, we are only considering the set of inertial measurements $\mathcal{I}_{0:k} \doteq \{\mathcal{I}_{0,1} \dots \mathcal{I}_{k-1,k}\}$. Thus, we can state a MAP estimation problem, where the posterior distribution to be maximized is:
    \begin{equation}
        p(\mathcal{Y}_k \vert \mathcal{I}_{0:k}) \propto p( \mathcal{I}_{0:k} \vert \mathcal{Y}_k) p(\mathcal{Y}_k)
    \end{equation}
    where $p( \mathcal{I}_{0:k} \vert \mathcal{Y}_k)$ stands for likelihood and $p(\mathcal{Y}_k)$ for prior. Considering independence of measurements, the inertial-only MAP estimation problem can be written as:
    \begin{equation}
        \mathcal{Y}_{k}^* = \underset{\mathcal{Y}_{k}}{\arg\max} \left(p(\mathcal{Y}_k) \prod_{i=1}^k p(\mathcal{I}_{i-1,i} \vert s, \textbf{R}_{\text{w}g}, \mathbf{b}, \mathbf{\bar{v}}_{i-1}, \mathbf{\bar{v}}_{i})  \right)
    \end{equation}
    
    Taking negative logarithm and assuming  Gaussian error for IMU preintegration and prior distribution, this finally results in the optimization problem:
    \begin{equation}
    \label{eq:imu_inertial_only}
        \mathcal{Y}_k^* = \underset{\mathcal{Y}_{k}}{\arg\min} \left( \Vert \mathbf{b}\Vert_{\Sigma^{-1}_b}^2 + \sum_{i=1}^{k} \Vert \mathbf{r}_{\mathcal{I}_{i-1,i}}\Vert_{\Sigma^{-1}_{\mathcal{I}_{i-1,i}}}^2 \right)   
    \end{equation}
    
    This optimization, represented in figure \ref{fig:Igraph}, differs from equation \ref{eq:imu_BA} in not including visual residuals, as the up-to-scale trajectory estimated by visual SLAM is taken as constant, and adding a prior residual that forces IMU biases to be close to zero. Covariance matrix $\Sigma_b$ represents prior knowledge about the range of values IMU biases may take. Details for preintegration of IMU covariance $\Sigma_{\mathcal{I}_{i-1,i}}$ can be found at \cite{forster2017manifold}. 
    
    As we are optimizing in a manifold we need to define a retraction \cite{forster2017manifold} to update $\textbf{R}_{ \text{wg}}$ during the optimization. Since rotation around gravity direction does not suppose a change in gravity, this update is parameterized with two angles $(\delta \alpha_{\textbf{g}}, \delta \beta_{\textbf{g}})$:
    \begin{equation}
    \textbf{R}_{ \text{wg}}^{\text{new}} = \textbf{R}_{ \text{wg}}^{\text{old}} \text{Exp}(\delta \alpha_{\textbf{g}}, \delta \beta_{\textbf{g}}, 0)
    \end{equation}
    
    \noindent being $\text{Exp}(.)$ the exponential map from $\mathbb{R}^3$ to $\text{SO}(3)$. To guarantee that scale factor remains positive during optimization we define its update as:
    \begin{equation}
    s^{\text{new}} = s^{\text{old}} \exp{(\delta s)}
    \end{equation}
    
    Once the inertial-only optimization is finished, the frame poses and velocities and the 3D map points are scaled with the estimated scale factor and rotated to align the $z$ axis with the estimated gravity direction. Biases are updated and IMU preintegration is repeated, aiming to reduce future linearization errors.

    \item {\bf Visual-Inertial MAP Estimation}: Once we have a good estimation for inertial and visual parameters, we can perform a joint visual-inertial optimization for further refining the solution. This optimization may be represented as figure \ref{fig:VIgraph} but having common biases for all keyframes and including the same prior information for biases than in the inertial-only step.
\end{enumerate}

Our exhaustive initialization experiments on the EuRoC dataset \cite{campos2020ICRA} show that this initialization is very efficient, achieving 5\% scale error with trajectories of 2 seconds. To improve the initial estimation, visual-inertial BA is performed 5 and 15 seconds after initialization, converging to 1\% scale error as shown in section \ref{sec:experiments}. After these BAs, we say that the map is {\em mature}, meaning that scale, IMU parameters and gravity directions are already accurately estimated. 

Our initialization is much more accurate than joint initialization methods that solve a set of algebraic equations \cite{martinelli2014closed,kaiser2017simultaneous, campos2019}, and much faster than the initialization used in ORB-SLAM-VI \cite{mur2017visual} that needed 15 seconds to get the first scale estimation, or that used in VI-DSO \cite{stumberg2018direct}, that starts with a huge scale error and requires 20-30 seconds to converge to 1\% error. Comparisons between different initialization methods may be found at \cite{campos2020ICRA}.

In some specific cases, when slow motion does not provide good observability of the inertial parameters, initialization may fail to converge to accurate solutions in just 15 seconds. To get robustness against this situation, we propose a novel scale refinement technique, based on a modified inertial-only optimization, where all inserted keyframes are included but scale and gravity direction are the only parameters to be estimated (figure \ref{fig:SGgraph}). Notice that in that case, the assumption of constant biases would not be correct. Instead, we use the values estimated from mapping, and we fix them.
This optimization, which is very computationally efficient, is performed in the Local Mapping thread every ten seconds, until the map has more than 100 keyframes, or more than 75 seconds have passed since initialization. 

Finally, we have easily extended our monocular-inertial initialization to stereo-inertial by fixing the scale factor to one and taking it out from the inertial-only optimization variables, enhancing its convergence.

\subsection{Tracking and Mapping}
For tracking and mapping we adopt the schemes proposed in \cite{mur2017visual}. Tracking solves a simplified visual-inertial optimization where only the states of the last two frames are optimized, while map points remain fixed. 

For mapping, trying to solve the whole optimization from equation \ref{eq:imu_BA} would be intractable for large maps. We use as optimizable variables a sliding window of keyframes and their points, including also observations to these points from covisible keyframes but keeping their pose fixed.

\subsection{Robustness to tracking loss}
In pure visual SLAM or VO systems, temporal camera occlusion and fast motions result in losing track of visual elements, getting the system lost. ORB-SLAM pioneered the use of fast relocalization techniques based on bag-of-words place recognition, but they proved insufficient to solve difficult sequences in the EuRoC dataset \cite{mur2017orb}. Our visual-inertial system enters into {\em visually lost} state when less than 15 point maps are tracked, and achieves robustness in two stages:
\begin{itemize}
    \item \textit{Short-term lost}: the current body state is estimated from IMU readings, and map points are projected in the estimated camera pose and searched for matches within a large image window. The resulting matches are included in visual-inertial optimization. In most cases this allows to recover visual tracking. Otherwise, after 5 seconds, we pass to the next stage.
    \item \textit{Long-term lost}: A new visual-inertial map is initialized as explained above, and it becomes the active map. 
\end{itemize}
If the system gets lost within 15 seconds after IMU initialization, the map is discarded. This prevents to accumulate inaccurate and meaningless maps.

\section{Map Merging and Loop Closing}

Short-term and mid-term data-associations between a frame and the active map are routinely found by the tracking and mapping threads by projecting map points into the estimated camera pose and searching for matches in an image window of just a few pixels. To achieve long-term data association for relocalization and loop detection, ORB-SLAM uses the DBoW2 bag-of-words place recognition system \cite{GalvezTRO12,mur2014ICRA}. This method has been also adopted by most recent VO and SLAM systems that implement loop closures (Table \ref{tab:related_work}). 

Unlike tracking, place recognition does not start from an initial guess for camera pose. Instead, DBoW2 builds a database of keyframes with their bag-of-words vectors, and given a query image is able to efficiently provide the most similar keyframes according to their bag-of-words. Using only the first candidate,  raw DBoW2 queries achieve precision and recall in the order of 50-80\% \cite{GalvezTRO12}. To avoid false positives that would corrupt the map, DBoW2 implements temporal and geometric consistency checks moving the working point to 100\% precision and 30-40\% recall \cite{GalvezTRO12,mur2014ICRA}. Crucially, the temporal consistency check delays place recognition at least during 3 keyframes. When trying to use it in our Atlas system, we found that this delay and the low recall resulted too often in duplicated areas in the same or in different maps. 

In this work we propose a new place recognition algorithm with improved recall for long-term and multi-map data association. Whenever the mapping thread creates a new keyframe, place recognition is launched trying to detect matches with any of the keyframes already in the Atlas. If the matching keyframe found belongs to the active map, a loop closure is performed. Otherwise, it is a multi-map data association, then, the active and the matching maps are merged. As a second novelty in our approach, once the relative pose between the new keyframe and the matching map is estimated, we define a {\em local window} with the matching keyframe and its neighbours in the covisibility graph. In this window we intensively search for mid-term data associations, improving the accuracy of loop closing and map merging. These two novelties explain the better accuracy obtained by ORB-SLAM3 compared with ORB-SLAM2 in the EuRoC experiments. The details of the different operations are explained next.

\subsection{Place Recognition}

To achieve higher recall, for every new active keyframe we query the DBoW2 database for several similar keyframes in the Atlas. To achieve 100\,\% precision, each of these candidates goes through several steps of geometric verification. The elementary operation of all the geometrical verification steps consists in checking whether there is an ORB keypoint inside an image window whose descriptor matches the ORB descriptor of a map point, using a threshold for the Hamming distance between them. If there are several candidates in the search window, to discard ambiguous matches, we check the distance ratio to the second-closest match \cite{lowe2004distinctive}. The steps of our place recognition algorithm are: 

\begin{enumerate}
    \item \textbf{DBoW2 candidate keyframes}. We query the Atlas DBoW2 database with the active keyframe $K_a$ to retrieve the three most similar keyframes, excluding keyframes covisible with $K_a$. We refer to each matching candidate for place recognition as $K_m$. 
    
    \item \textbf{Local window}.
    For each $K_m$ we define a local window that includes $K_m$, its best covisible keyframes, and the map points observed by all of them. The DBoW2 direct index provides a set of putative matches between keypoints in $K_a$ and in the local window keyframes. For each of these 2D-2D matches we have also available the 3D-3D match between their corresponding map points.
    
    \item \textbf{3D aligning transformation}. We compute using RANSAC the transformation $\mathbf{T}_{am}$  that better aligns the map points in $K_m$ local window with those of $K_a$. In pure monocular, or in monocular-inertial when the map is still not mature, we compute $\mathbf{T}_{am}\in \text{Sim}(3)$, otherwise $\mathbf{T}_{am}\in \text{SE}(3)$. In both cases we use Horn algorithm \cite{horn1987closed} using a minimal set of three 3D-3D matches to find each hypothesis for $\mathbf{T}_{am}$. The putative matches that, after transforming the map point in $K_a$ by $\mathbf{T}_{am}$, achieve a reprojection error in $K_a$ below a threshold, give a positive vote to the hypothesis. The hypothesis with more votes is selected, provided the number is over a threshold.
    
    \item \textbf{Guided matching refinement}. All the map points in the local window are transformed with $\mathbf{T}_{am}$ to find more matches with the keypoints in $K_a$. The search is also reversed, finding matches for $K_a$ map points in all the keyframes of the local window. Using all the matchings found, $\mathbf{T}_{am}$ is refined by non-linear optimization, where the goal function is the bidirectional reprojection error, using Huber influence function to provide robustness to spurious matches. 
    If the number of inliers after the optimization is over a threshold, a second iteration of guided matching and non-linear refinement is launched, using a smaller image search window. 
    
    \item \textbf{Verification in three covisible keyframes}.  To avoid false positives, DBoW2 waited for place recognition to fire in three consecutive keyframes, delaying or missing place recognition. Our crucial insight is that, most of the time, the information required for verification is already in the map. To verify place recognition, we search in the active part of the map two keyframes covisible with $K_a$ where the number of matches with points in the local window is over a threshold. If they are not found, the validation is further tried with the new incoming keyframes, without requiring the bag-of-words to fire again. The validation continues until three keyframes verify $\mathbf{T}_{am}$, or two consecutive new keyframes fail to verify it.
    
    \item \textbf{VI Gravity direction verification}. In the visual-inertial case, if the active map is mature, we have estimated $\mathbf{T}_{am}\in \text{SE}(3)$. We further check whether the pitch and roll angles are below a threshold to definitively accept the place recognition hypothesis.
\end{enumerate}

\subsection{Visual Map Merging}
When a successful place recognition produces multi-map data association between  keyframe $K_a$ in the active map $M_a$, and a matching keyframe $K_m$ from a different map stored in the Atlas $M_m$, with an aligning transformation $\mathbf{T}_{am}$, we launch a map merging operation. In the process, special care must be taken to ensure that the information in $M_m$ can be promptly reused by the tracking thread to avoid map duplication.
For this, we propose to bring the $M_a$ map into $M_m$ reference. As $M_a$ may contain many elements and merging them might take a long time, merging is split in two steps. First, the merge is performed in a {\em welding window} defined by the neighbours of $K_a$ and $K_m$ in the covisibility graph, and in a second stage, the correction is propagated to the rest of the merged map by a pose-graph optimization. The detailed steps of the merging algorithm are:
\begin{enumerate}
    \item \textbf{Welding window assembly}. The welding window includes $K_a$ and its covisible keyframes, $K_m$ and its covisible keyframes, and all the map point observed by them. Before their inclusion in the welding window, the keyframes and map points belonging to $M_a$ are transformed by $\mathbf{T}_{ma}$ to align them with respect to $M_m$.
 
    \item \textbf{Merging maps}. Maps $M_a$ and $M_m$ are fused together to become the new active map. To remove duplicated points, matches are actively searched for $M_a$ points in the $M_m$ keyframes. For each match, the point from $M_a$ is removed, and the point in $M_m$ is kept accumulating all the observations of the removed point. The covisibility and essential graphs \cite{mur2015orb} are updated by the addition of edges connecting keyframes from $M_m$ and $M_a$ thanks to the new mid-term point associations found.
    
    \item \textbf{Welding bundle adjustment}. A local BA is performed optimizing all the keyframes from $M_a$ and $M_m$ in the welding window along with the map points which are observed by them (Fig.\,\ref{fig:merge_visual_BA_sub}). To fix gauge freedom, the keyframes of $M_m$ not belonging to the welding window but observing any of the local map points are included in the BA with their poses fixed. Once the optimization finishes, all the keyframes included in the welding area can be used for camera tracking, achieving fast and accurate reuse of map $M_m$.
    
    \item \textbf{Essential-graph optimization}. A pose-graph optimization is performed using the essential graph of the whole merged map, keeping fixed the keyframes in the welding area. This optimization propagates corrections from the welding window to the rest of the map. 
    
\end{enumerate}

\subsection{Visual-Inertial Map Merging}
\begin{figure}
    \begin{subfigure}{\columnwidth}
      \includegraphics[width=1.0\columnwidth]{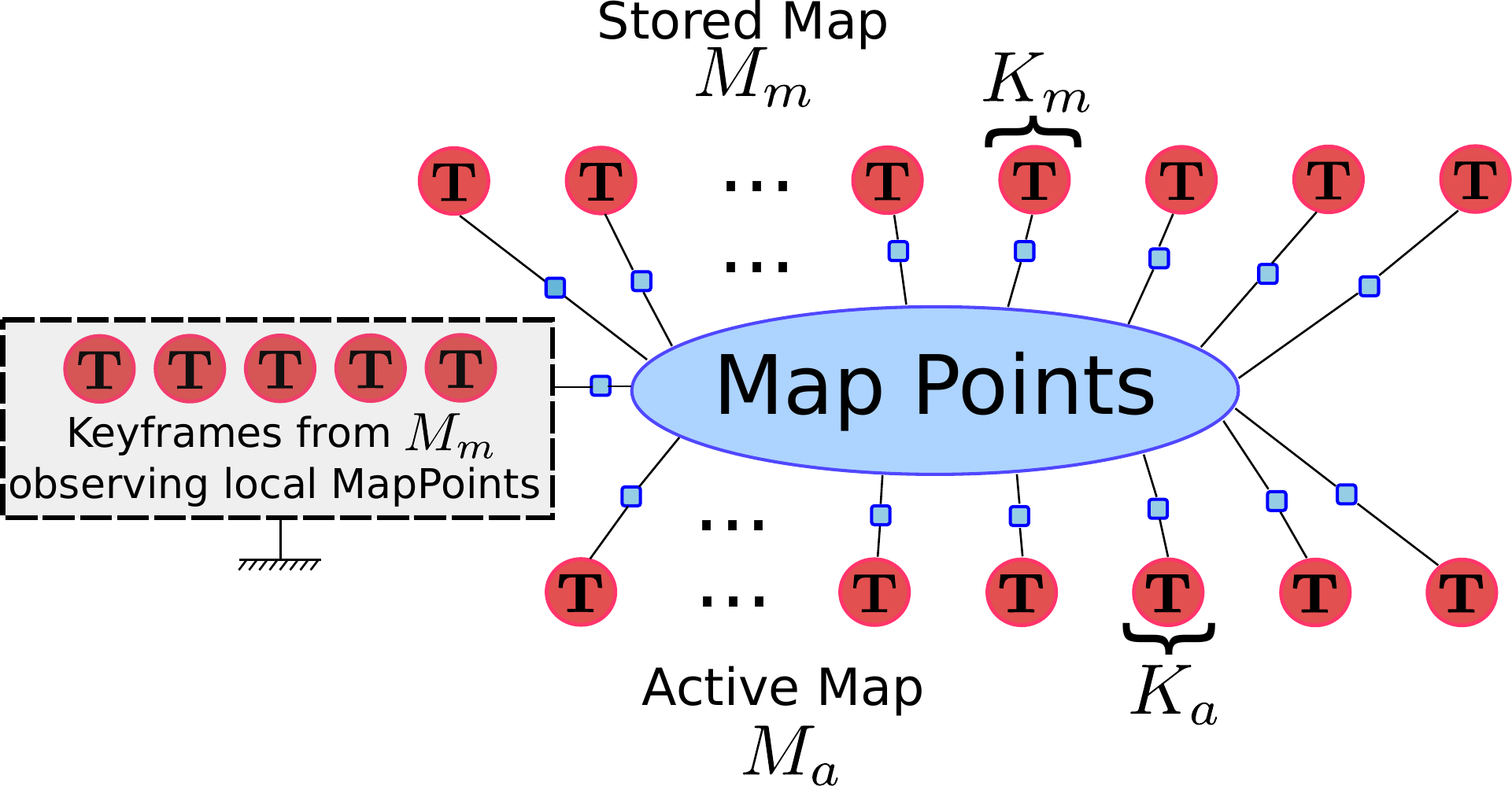}
      \caption{Visual welding BA}
      \label{fig:merge_visual_BA_sub}
    \end{subfigure}
    \begin{subfigure}{\columnwidth}
      \includegraphics[width=1.0\columnwidth]{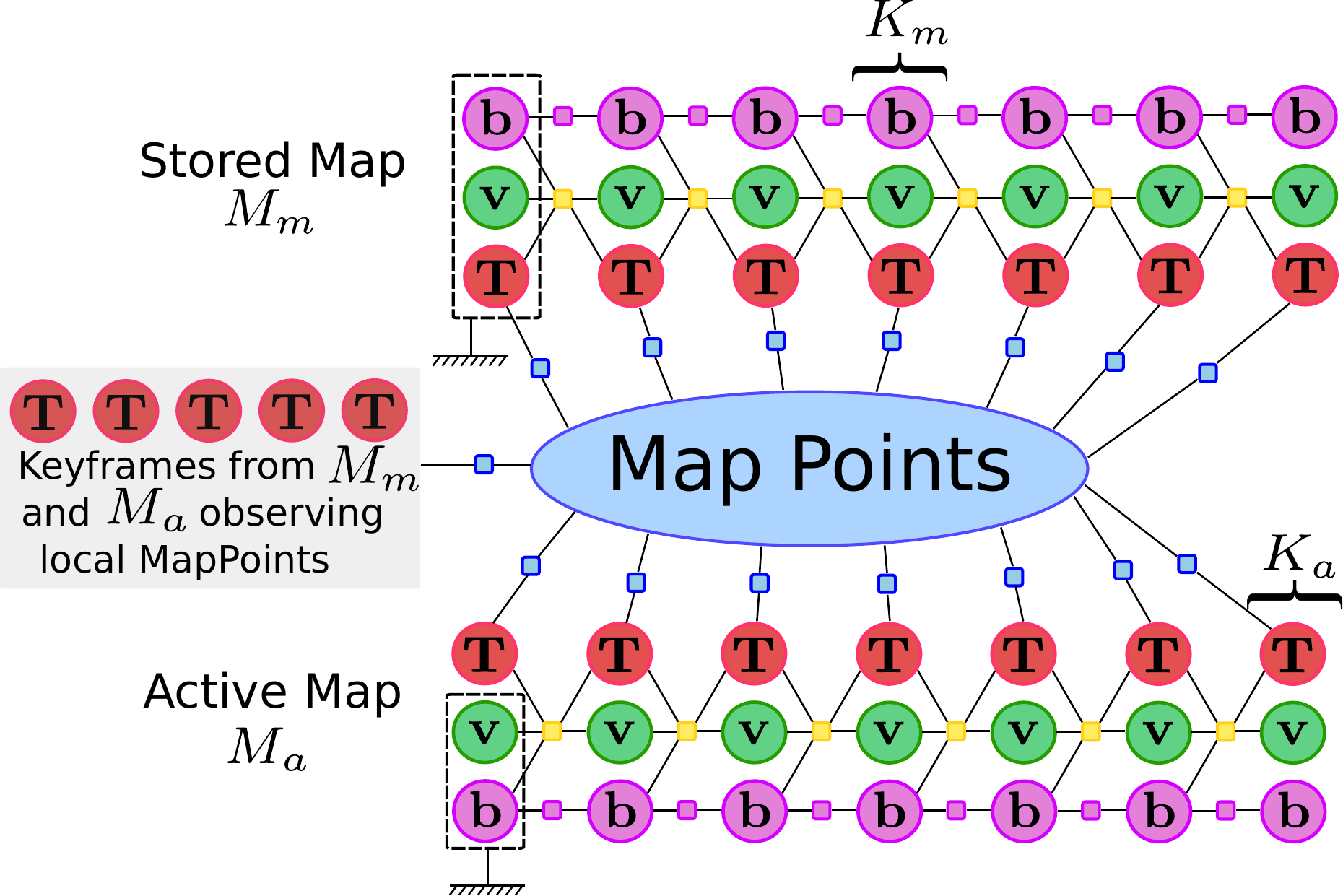}
      \caption{Visual-Inertial welding BA}
       \label{fig:merge_VI_BA_sub}
    \end{subfigure}
      \caption{Factor graph representation for the welding BA, with reprojection error terms (blue squares), IMU preintegration terms (yellow squares) and bias random walk (purple squares).}
 
      \label{fig:welding_BA}
    \end{figure}

The visual-inertial merging algorithm follows similar steps than the pure visual case. Steps 1) and 3) are modified to better exploit the inertial information:
\begin{enumerate}
    \item \textbf{VI welding window assembly}: If the active map is mature, we apply the available $\mathbf{T}_{ma}\in \text{SE}(3)$ to map $M_a$ before its inclusion in the welding window. If the active map is not mature, we align $M_a$ using the available $\mathbf{T}_{ma}\in \text{Sim}(3)$.
    
    \item \textbf{VI welding bundle adjustment}: Poses, velocities and biases of keyframes $K_a$ and $K_m$ and their five last temporal keyframes are included as optimizable. These variables are related by IMU preintegration terms, as shown in Figure \ref{fig:merge_VI_BA_sub}. For $M_m$, the keyframe immediately before the local window is included but fixed, while for $M_a$ the similar keyframe is included but its pose remains optimizable. All map points seen by the above mentioned keyframes are optimized, together with poses from $K_m$ and $K_a$ covisible keyframes. All keyframes and points are related by means of reprojection error.
\end{enumerate}
    
\subsection{Loop Closing}
Loop closing correction algorithm is analogous to map
merging, but in a situation where both keyframes matched
by place recognition belong to the active map. A welding
window is assembled from the matched keyframes, and point
duplicates are detected and fused creating new links in the
covisibility and essential graphs. The next step is a pose-graph optimization to propagate the loop correction to the rest of the map. The final step is a global BA to find the MAP estimate after considering the loop closure mid-term and long-term matches. In the visual-inertial case, the global BA is only performed if the number of keyframes is below a threshold to avoid a huge computational cost.


\begin{table*} 
\centering
\scriptsize
\caption{Performance comparison in the EuRoC dataset (RMS ATE in m., scale error in \%). Except where noted, we show results reported by the authors of each system, for all the frames in the trajectory, comparing with the processed GT. }
\label{table:euro_sensor_error}
\begin{tabular}{|c|c @{\hspace{2.5ex}} c|c @{\hspace{2.5ex}} c @{\hspace{2.5ex}} c c @{\hspace{2.5ex}} c| @{\hspace{2.5ex}}c c @{\hspace{2.5ex}} c| @{\hspace{2.5ex}} c @{\hspace{2.5ex}} c @{\hspace{2.5ex}} c|c|} \hline
                  & & & MH01 & MH02 & MH03 & MH04 & MH05 & V101 & V102 & V103 & V201 & V202 & V203 & Avg$^1$ \\ 
\hline \hline
\multirow{9}{*}{\begin{tabular}[c]{@{}c@{}} Monocular \end{tabular}} 
& \doble{ORB-SLAM}{\cite{mur2017visual}} & ATE$^{2,3}$  & 0.071 & 0.067 & 0.071 & 0.082 & \textbf{0.060} & \textbf{0.015} & 0.020 & - & \textbf{0.021} & \textbf{0.018} & - & 0.047* \\
\cline{2-15}
 & \doble{DSO}{\cite{Engel-et-al-pami2018}} & ATE  & 0.046 & 0.046 & 0.172 & 3.810 & 0.110 & 0.089 & 0.107 & 0.903 & 0.044 & 0.132 & 1.152 & 0.601 \\
\cline{2-15}
 & \doble{SVO}{\cite{forster2017svo}} & ATE  & 0.100 & 0.120 & 0.410 & 0.430 & 0.300 & 0.070 & 0.210 & - & 0.110 & 0.110 & 1.080 & 0.294* \\ 
\cline{2-15}
& \doble{DSM}{\cite{zubizarreta2020direct}} & ATE  & 0.039 & 0.036 & 0.055 & \textbf{0.057} & 0.067 & 0.095 & 0.059 & 0.076 & 0.056 & 0.057 & \textbf{0.784} & \textbf{0.126} \\
\cline{2-15}
 & \doble{ORB-SLAM3}{(ours)} & ATE  & \textbf{0.016} & \textbf{0.027} & \textbf{0.028} & 0.138 & 0.072 & 0.033 & \textbf{0.015} & \textbf{0.033} & 0.023 & 0.029 & - & 0.041* \\  
 \hline \hline
\multirow{7}{*}{Stereo} & 
\doble{ORB-SLAM2}{\cite{mur2017orb}} & ATE  & 0.035 & \textbf{0.018} & 0.028 & 0.119 & 0.060 & \textbf{0.035} & \textbf{0.020} & \textbf{0.048} & \textbf{0.037} & 0.035 & - & 0.044* \\
\cline{2-15}
 & \doble{VINS-Fusion}{\cite{qin2019general}} & ATE  & 0.540 & 0.460 & 0.330 & 0.780 & 0.500 & 0.550 & 0.230 & - & 0.230 & 0.200 & - & 0.424* \\
\cline{2-15}
 & \doble{SVO}{\cite{forster2017svo}} & ATE  & 0.040 & 0.070 & 0.270 & 0.170 & 0.120 & 0.040 & 0.040 & 0.070 & 0.050 & 0.090 & 0.790 & 0.159 \\ 
\cline{2-15}
 & \begin{tabular}[c]{@{}c@{}} ORB-SLAM3 \\ (ours) \end{tabular} & ATE & \textbf{0.029} & 0.019 & \textbf{0.024} & \textbf{0.085} & \textbf{0.052} & \textbf{0.035} & 0.025 & 0.061 & 0.041 & \textbf{0.028} & \textbf{0.521} & \textbf{0.084} \\
\hline \hline
\multirow{13}{*}{\begin{tabular}[c]{@{}c@{}} Monocular \\ Inertial \end{tabular}}
& \doble{MCSKF}{\cite{mourikis2007multi}} & ATE$^5$ & 0.420 & 0.450 & 0.230 & 0.370 & 0.480 & 0.340 & 0.200 & 0.670 & 0.100 & 0.160 & 1.130 & 0.414\\
\cline{2-15}
& \doble{OKVIS}{ \cite{leutenegger2015keyframe}} & ATE$^5$ & 0.160 & 0.220 & 0.240 & 0.340 & 0.470 & 0.090 & 0.200 & 0.240 & 0.130 & 0.160 & 0.290 & 0.231\\
\cline{2-15}
& \doble{ROVIO}{\cite{bloesch2017iterated}} & ATE$^5$ & 0.210 & 0.250 & 0.250 & 0.490 & 0.520 & 0.100 & 0.100 & 0.140 & 0.120 & 0.140 & 0.140 & 0.224\\

\cline{2-15}
& \multirow{2}{*}{\begin{tabular}[c]{@{}c@{}} ORBSLAM-VI \\ \cite{mur2017visual} \end{tabular}} & ATE$^{2,3}$ & 0.075 & 0.084 & 0.087 & 0.217 & 0.082 & \textbf{0.027} & 0.028 & - & \textbf{0.032} & 0.041 & 0.074 & 0.075* \\
& & scale error$^{2,3}$ & 0.5 & 0.8 & 1.5 & 3.5 & 0.5 & 0.9 & 0.8 & -& 0.2 & 1.4 & 0.7 & 1.1* \\
\cline{2-15}
& \doble{VINS-Mono}{\cite{qin2018vins}}  & ATE$^4$ & 0.084 & 0.105 & 0.074 & 0.122 & 0.147 & 0.047 & 0.066 & 0.180 & 0.056 & 0.090 & 0.244 & 0.110 \\
\cline{2-15}
& \multirow{2}{*}{\begin{tabular}[c]{@{}c@{}} VI-DSO \\ \cite{stumberg2018direct} \end{tabular}} & ATE  & \textbf{0.062} & 0.044 & 0.117 & 0.132 & 0.121 & 0.059 & 0.067 & 0.096 & 0.040 & 0.062 & 0.174 & 0.089 \\
& & scale error & 1.1 & 0.5 & 0.4 & 0.2 & 0.8 & 1.1 & 1.1 & 0.8 & 1.2 & 0.3 & 0.4 & 0.7 \\
 \cline{2-15}
 & \multirow{2}{*}{\begin{tabular}[c]{@{}c@{}} ORB-SLAM3 \\ (ours) \end{tabular}} & ATE & \textbf{0.062} & \textbf{0.037} & \textbf{0.046} & \textbf{0.075} & \textbf{0.057} & 0.049 & \textbf{0.015} & \textbf{0.037} & 0.042 & \textbf{0.021} & \textbf{0.027} & \textbf{0.043} \\
 & & scale error & 1.4 & 0.3 & 0.8 & 0.5 & 0.3 & 2.0 & 0.6 & 2.2 & 0.7 & 0.4 & 1.0 & 0.9 \\
\hline \hline

\multirow{7}{*}{\begin{tabular}[c]{@{}c@{}} Stereo \\ Inertial \end{tabular}} 
& \doble{VINS-Fusion}{\cite{qin2019general}} & ATE$^4$ & 0.166 & 0.152 & 0.125 & 0.280 & 0.284 & 0.076 & 0.069 & 0.114 & 0.066 & 0.091 & 0.096 & 0.138 \\
\cline{2-15}
 & \begin{tabular}[c]{@{}c@{}} BASALT \\ \cite{usenko2020visual} \end{tabular}   & ATE$^3$ & 0.080 & 0.060 & 0.050 & 0.100 & \textbf{0.080} & 0.040 & 0.020 & 0.030 & \textbf{0.030} & 0.020 & - & 0.051* \\
\cline{2-15}
 & \doble{Kimera}{ \cite{Rosinol20icra-Kimera}}   & ATE & 0.080 & 0.090 & 0.110 & 0.150 & 0.240 & 0.050 & 0.110 & 0.120 & 0.070 & 0.100 & 0.190 & 0.119 \\
\cline{2-15}
& \multirow{2}{*}{\begin{tabular}[c]{@{}c@{}} ORB-SLAM3 \\ (ours) \end{tabular}} & ATE & \textbf{0.036} & \textbf{0.033} & \textbf{0.035} & \textbf{0.051} & 0.082 & \textbf{0.038} & \textbf{0.014} & \textbf{0.024} & 0.032 & \textbf{0.014} & \textbf{0.024} & \textbf{0.035} \\
 & & scale error & 0.6 & 0.2 & 0.6 & 0.2 & 0.9 & 0.8 & 0.6 & 0.8 & 1.1 & 0.2 & 0.2 & 0.6 \\
\hline
\multicolumn{15}{l}{\begin{tabular}[l]{@{}l@{}} \\ 

$^1$ Average error of the successful sequences. Systems that did no complete all sequences are denoted by * and are not marked in bold. \\
$^2$ Errors reported with raw GT instead of processed GT. \\ 
$^3$ Errors reported with keyframe trajectory instead of full trajectory. \\
$^4$ Errors obtained by ourselves, running the code with its default configuration.\\
$^5$ Errors reported at \cite{delmerico2018benchmark}.\\
\end{tabular} }\\
\end{tabular}
\end{table*}

\section{Experimental Results}
\label{sec:experiments}
The evaluation of the whole system is split in:
\begin{itemize}
    \item Single session experiments in EuRoC \cite{burri2016euroc}: each of the 11 sequences is processed to produce a map, with the four sensor configurations: Monocular, Monocular-Inertial, Stereo and Stereo-Inertial.
    \item Performance of monocular and stereo visual-inertial SLAM with fisheye cameras, in the challenging TUM VI Benchmark \cite{schubert2018tum}.
    \item Multi-session experiments in both datasets.

\end{itemize}
As usual in the field, we measure accuracy with RMS ATE \cite{sturm2012benchmark}, aligning the estimated trajectory with ground-truth using a $\mbox{Sim}(3)$ transformation in the pure monocular case, and a $\mbox{SE}(3)$ transformation in the rest of sensor configurations. 
Scale error is computed using $s$ from  $\mbox{Sim}(3)$ alignment, as $\lvert 1-s \rvert$.
All experiments have been run on an Intel Core i7-7700 CPU, at 3.6GHz, with 32 GB memory, using only CPU.

\subsection{Single-session SLAM on EuRoC}


\begin{figure}
  \includegraphics[width=\columnwidth]{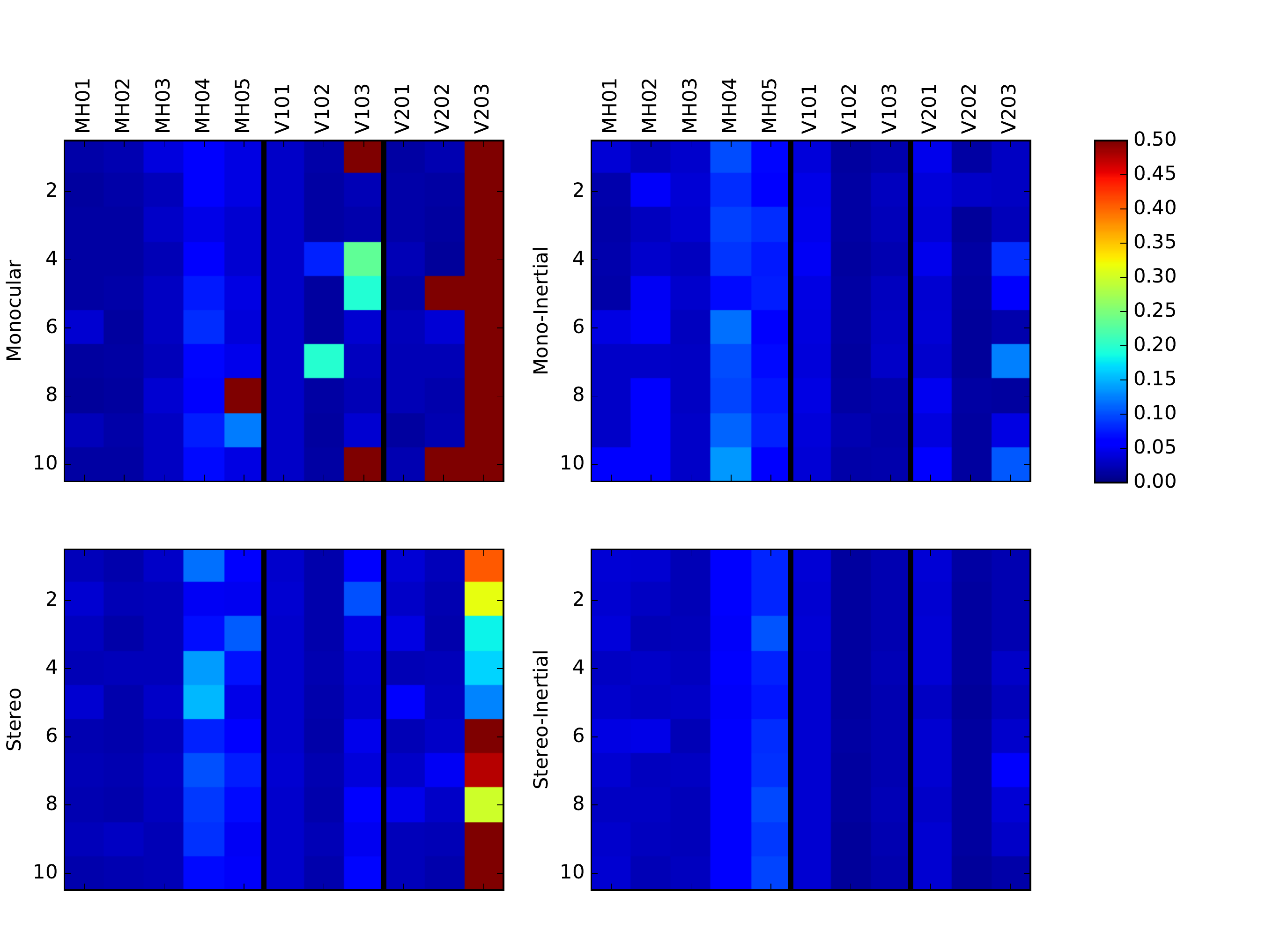}
  \caption{Colored squares represent the RMS ATE for ten different execution in each sequence of the EuRoC dataset.} 
  \label{fig:error_euroc_executions}
\end{figure}


Table \ref{table:euro_sensor_error} compares the performance of ORB-SLAM3 using its four sensor configurations with the most relevant systems in the state-of-the-art. Our reported values are the median after 10 executions. As shown in the table, ORB-SLAM3 achieves in all sensor configurations more accurate result than the best systems available in the literature, in most cases by a wide margin. 

In monocular and stereo configurations our system is more precise than ORB-SLAM2 due to the better place recognition algorithm that closes loops earlier and provides more mid-term matches. Interestingly, the next best results are obtained by DSM that also uses mid-term matches, even though it does not close loops. 

In monocular-inertial configuration, ORB-SLAM3  is five to ten times more accurate than MCSKF, OKVIS and ROVIO, and more than doubles the accuracy of VI-DSO and VINS-Mono, showing again the advantages of mid-term and long-term data association. Compared with ORB-SLAM VI, our novel fast IMU initialization allows ORB-SLAM3 to calibrate the inertial sensor in a few seconds and use it from the very beginning, being able to complete all EuRoC sequences, and obtaining better accuracy.

In stereo-inertial configuration, ORB-SLAM3 is three to four times more accurate than and Kimera and VINS-Fusion. It's accuracy is only approached by the recent BASALT that, being a native stereo-inertial system, was not able to complete sequence V203, where some frames from one of the cameras are missing. Comparing our monocular-inertial and stereo-inertial systems, the latter performs better in most cases. Only for two Machine Hall (MH) sequences a lower accuracy is obtained. We hypothesize that greater depth scene for MH sequences may lead to less accurate stereo triangulation and hence a less precise scale.

To summarize performance, we have presented the median of ten executions for each sensor configuration. For a robust system, the median represents accurately the behavior of the system. But a non-robust system will show high variance in its results. This can be analyzed using figure \ref{fig:error_euroc_executions} that shows with colors the error obtained in each of the ten executions. Comparison with the figures for DSO, ROVIO and VI-DSO published in \cite{stumberg2018direct}  confirms the superiority of our method. 

In pure visual configurations, the multi-map system adds some robustness to fast motions by creating a new map when tracking is lost, that is merged later with the global map. This can be seen in sequences V103 monocular and V203 stereo that could not be solved by ORB-SLAM2 and are successfully solved by our system in most executions. As expected, stereo is more robust than monocular thanks to its faster feature initialization, with the additional advantage that the real scale is estimated.

However, the big leap in robustness is obtained by our novel visual-inertial SLAM system, both in monocular and stereo configurations. The stereo-inertial system has a very slight advantage over monocular-inertial, particularly in the most challenging V203 sequence.

We can conclude that inertial integration not only boosts accuracy, reducing the median ATE error compared to pure visual solutions, but it also endows the system with excellent robustness, having a much more stable performance.

\subsection{Visual-Inertial SLAM on TUM-VI Benchmark}
The TUM-VI dataset \cite{schubert2018tum} consists of 28 sequences in 6 different environments, recorded using a hand-held fisheye stereo-inertial rig. Ground-truth for the trajectory is only available at the beginning and at the end of the sequences, which for most of them represents a very small portion of the whole trajectory. Many sequences in the dataset do not contain loops. Even if the starting and ending point are in the same room, point of view directions are opposite and place recognition cannot detect any common region. Using this ground-truth for evaluation amounts to measuring the accumulated drift along the whole trajectory. 

We extract 1500 ORB points per image in monocular-inertial setup, and 1000 points per image in stereo-inertial, after applying CLAHE equalization to address under and over exposure found in the dataset.
For outdoors sequences, our system struggles with very far points coming from the cloudy sky, that is very visible in fisheye cameras. These points may have slow motion that can  introduce drift in the camera pose. For preventing this, we discard points further than 20 meters from the current camera pose, only for outdoors sequences. A more sophisticated solution would be to use an image segmentation algorithm to detect and discard the sky. 

The results obtained are compared with the most relevant systems in the literature in table \ref{table:tum_vi}, that clearly shows the superiority of ORB-SLAM3 both in monocular-inertial and stereo-inertial. The closest systems are VINS-Mono and BASALT, that are essentially visual-inertial odometry systems with loop closures, and miss mid-term data associations.

Analyzing more in detail the performance of our system, it gets lowest error in small and medium indoor environments, {\em room} and {\em corridor} sequences, with errors below 10\,cm for most of them. In these trajectories, the system is continuously revisiting and reusing previously mapped regions, which is one of the main strengths of ORB-SLAM3. Also, tracked points are typically closer than 5\,m, what makes easier to estimate inertial parameters, preventing them from diverging.

In {\em magistrale} indoors sequences, that are up to 900\,m long, most tracked points are relatively close, and ORB-SLAM3 obtains errors around 1\,m except in one sequence that goes close to 5\,m. In contrast, in some long {\em outdoors} sequences, the scarcity of close visual features may cause drift of the inertial parameters, notably scale and accelerometer bias, which leads to errors in the order of 10 to 70 meters. Even though, ORB-SLAM3 is the best performing system in the outdoor sequences.

This dataset also contains three really challenging {\em slides} sequences, where the user descends though a dark tubular slide with almost total lack of visual features. In this situation, a pure visual system would be lost, but our visual-inertial system is able to process the whole sequence with competitive error, even if no loop-closures can be detected. Interestingly, VINS-Mono and BASALT, that track features using Lukas-Kanade, obtain in some of these sequences better accuracy than ORB-SLAM3, that matches ORB descriptors.

\setlength\tabcolsep{1.5pt}
\begin{table}
\centering
\scriptsize
\caption{TUM VI Benchmark \cite{schubert2018tum}: RMS ATE (m) for regions with available ground-truth data.}
\begin{tabular}{|   c   ||   c  |   c   ||   c  |   c |  c  |  c  || c  |  c  |}
\hline
&   \multicolumn{2}{   c  ||}{Mono-Inertial} & \multicolumn{4}{   c  ||}{Stereo-Inertial} & & \\
\cline{2-7}
Seq. & \doble{VINS-}{Mono} & \doble{ORB-}{SLAM3} & OKVIS & ROVIO & BASALT  & \doble{ORB-}{SLAM3}  & \doble{Length}{(m)} & LC \\
\hline
corridor1 & 0.63 & \textbf{0.04} & 0.33 & 0.47 & 0.34 & \textbf{0.03} & 305 & \checkmark \\
corridor2 & 0.95  & \textbf{0.02} & 0.47 & 0.75 & 0.42 & \textbf{0.02} &  322 & \checkmark \\
corridor3 & 1.56  & \textbf{0.31} & 0.57 & 0.85 & 0.35 & \textbf{0.02} &  300 & \checkmark \\
corridor4 & 0.25  & \textbf{0.17} & 0.26 & \textbf{0.13} & 0.21 & 0.21  & 114 & \\ 
corridor5 & 0.77 &  \textbf{0.03} & 0.39 & 2.09 & 0.37 & \textbf{0.01} &  270 & \checkmark \\
\hline
magistrale1 & 2.19 & \textbf{0.56} & 3.49 & 4.52 & 1.20 & \textbf{0.24} &  918 & \checkmark \\
magistrale2 & 3.11 &  \textbf{0.52} & 2.73 & 13.43 & 1.11 & \textbf{0.52} &   561 & \checkmark \\
magistrale3 & \textbf{0.40} & 4.89 & 1.22 & 14.80 & \textbf{0.74} & 1.86  & 566 &  \\
magistrale4 & 5.12 & \textbf{0.13} & {0.77} & 39.73 & 1.58 & \textbf{0.16} &  688 & \checkmark \\
magistrale5 &  \textbf{0.85}  & 1.03 & 1.62 & 3.47 & \textbf{0.60} & 1.13  & 458 & \checkmark \\ 
magistrale6 & 2.29 & \textbf{1.30} & 3.91 & X & 3.23 & \textbf{0.97}  & 771 &  \\ 
\hline
outdoors1 &  74.96 & \textbf{70.79} & X & 101.95 & 255.04 & \textbf{32.23} & 2656 &  \\
outdoors2 &  133.46 & \textbf{14.98} & 73.86 & 21.67 & 64.61 & \textbf{10.42}  & 1601 &  \\
outdoors3 & \textbf{36.99} & 39.63* & 32.38 & \textbf{26.10} & 38.26 & 54.77  & 1531 &  \\
outdoors4 &  \textbf{16.46} & 25.26 & 19.51 & X & 17.53 & \textbf{11.61}  & 928 &  \\
outdoors5 & 130.63 & \textbf{14.87} & {13.12} & 54.32 & \textbf{7.89} & 8.95 &  1168 & \checkmark \\
outdoors6 & 133.60 & \textbf{16.84} & 96.51 & 149.14 & 65.50 & \textbf{10.70}  & 2045 &  \\
outdoors7 & 21.90 & \textbf{7.59} & 13.61 & 49.01 & \textbf{4.07} & 4.58 &  1748 & \checkmark \\ 
outdoors8 & 83.36 & \textbf{27.88} & 16.31 & 36.03 & 13.53 & \textbf{11.02} & 986 &  \\
\hline
room1 & 0.07  & \textbf{0.01} & 0.06 & 0.16 & 0.09 & \textbf{0.01}  & 146 & \checkmark  \\
room2 & 0.07  & \textbf{0.02} & 0.11 & 0.33 & 0.07 & \textbf{0.01}  & 142 & \checkmark \\
room3 & 0.11  & \textbf{0.04} & 0.07 & 0.15 & 0.13 & \textbf{0.01}  & 135 & \checkmark \\
room4 & 0.04  & \textbf{0.01} & 0.03 & 0.09 & 0.05 & \textbf{0.01}  & 68 & \checkmark \\
room5 & 0.20  & \textbf{0.02} & 0.07 & 0.12 & 0.13 & \textbf{0.01}  & 131 & \checkmark \\
room6 & 0.08  & \textbf{0.01} & 0.04 & 0.05 & 0.02 & \textbf{0.01}  & 67 & \checkmark \\
\hline
slides1 & \textbf{0.68} & 0.97 & 0.86 & 13.73 & \textbf{0.32} & 0.41  & 289 &  \\
slides2 & \textbf{0.84} & 1.06 & 2.15 & 0.81 & \textbf{0.32} & 0.49  & 299 &  \\ 
slides3 & \textbf{0.69} & \textbf{0.69} & 2.58 & 4.68 & 0.89 & \textbf{0.47}  & 383 & \\
\hline

\multicolumn{7}{l}{ \begin{tabular}[l]{@{}l@{}} Ours are median of three executions.\\
For other systems, we provide values reported at \cite{schubert2020tum} \\
* points out that one out of three runs has not been successful \\
LC: Loop Closing may exist in that sequence
\end{tabular}} \\

\end{tabular}
\label{table:tum_vi}
\end{table}
\setlength\tabcolsep{6pt}

\begin{table}
\centering
\caption{RMS ATE (m) obtained by ORB-SLAM3 with four sensor configurations  in the room sequences, representative of AR/VR scenarios (median of 3 executions).}
\begin{tabular}{|  c   ||  c  |  c  |  c  |  c  |}
\hline
Seq. & Mono & Stereo & \doble{Mono-}{Inertial} & \doble{Stereo-}{Inertial} \\
\hline
room1 & {0.042} & {0.077} & 0.009 & 0.008 \\
room2 & {0.026} & 0.055 & 0.018 & 0.012 \\
room3 & {0.028} & 0.076 & 0.008 & 0.011 \\
room4 & {0.046} & {0.071} & 0.009 & 0.008 \\
room5 & {0.046} & {0.066} & 0.014 & 0.010 \\
room6 & 0.043 & {0.063} &  0.006 & 0.006 \\
\hline
Avg. & 0.039 & 0.068 & 0.011 & 0.009 \\
\hline
\end{tabular}
\label{table:tum_vi_arvr}
\end{table}

\setlength\tabcolsep{3.0pt}
\begin{table}
\centering
\caption{Multi-session RMS ATE (m) on the EuRoC dataset. For CCM-SLAM and VINS we show results reported by the authors of each system. Our values are the median of 5 executions, aligning the trajectories with the processed GT.}
\begin{tabular}{|l c|c|c|c|c|} \hline
Room & & \multicolumn{2}{c|}{Machine Hall} & Vicon 1 & Vicon 2 \\ \hline
Sequences & & \begin{tabular}[c]{@{}c@{}}MH01-03\end{tabular} & \begin{tabular}[c]{@{}c@{}}MH01-05\end{tabular} & V101-103 & V201-203 \\
\hline \hline
\begin{tabular}[c]{@{}c@{}} ORB-SLAM3 \\ Mono \end{tabular} & ATE & 0.030 & 0.058 & 0.058  & 0.284  \\ \hline
\begin{tabular}[c]{@{}c@{}} CCM-SLAM \\ Mono\cite{schmuck2018ccm} \end{tabular} & ATE & 0.077 & - & - & - \\
\hline \hline
\begin{tabular}[c]{@{}c@{}} ORB-SLAM3 \\ Stereo \end{tabular} & ATE & 0.028 & 0.040 & 0.027 & 0.163 \\
\hline \hline
\multirow{2}{*}{\begin{tabular}[c]{@{}c@{}} ORB-SLAM3 \\ Mono-Inertial \end{tabular}} & ATE & 0.037 & 0.065 & 0.040 & 0.048 \\
 & Scale error & 0.4 & 0.3 & 1.4 & 0.9 \\
\hline
\begin{tabular}[c]{@{}c@{}} VINS\cite{qin2018vins} \\ Mono-Inertial \end{tabular} & ATE & - & 0.210 & - & -  \\
\hline
\hline
\multirow{2}{*}{\begin{tabular}[c]{@{}c@{}} ORB-SLAM3 \\ Stereo-Inertial \end{tabular}} & ATE & 0.041 & 0.047 & 0.031 & 0.046 \\ 
 & Scale error & 0.6 & 0.3 & 0.6 & 0.8 \\
\hline
\end{tabular}
\label{table:multiple_session_experiment}
\end{table}
\setlength\tabcolsep{6pt}

\begin{figure*}
\centering
  \includegraphics[width=2.\columnwidth]{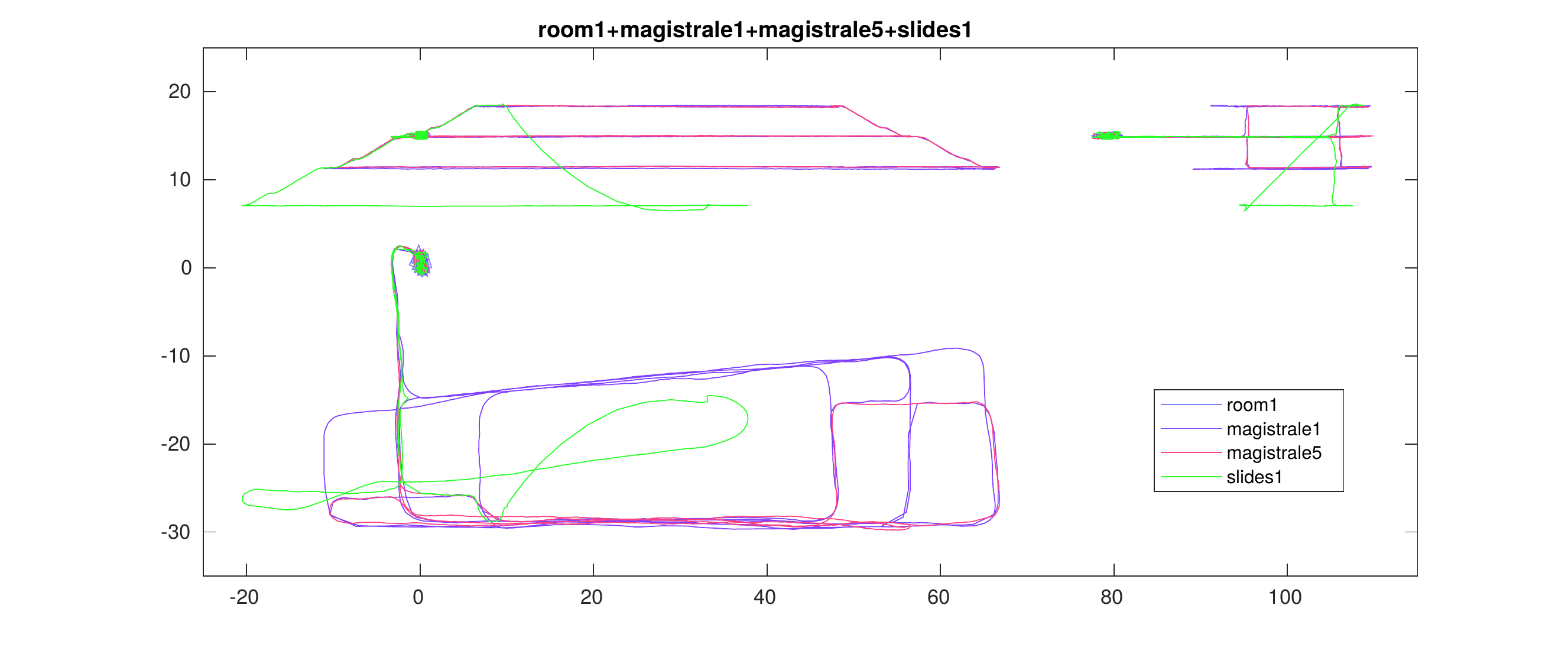}
  \caption{Multi-session stereo-inertial result with several sequences from TUM-VI dataset (front, side and top views).}
  \label{fig:tum_map}
\end{figure*}

Finally, the {\em room} sequences can be representative of typical AR/VR applications, where the user moves with a hand-held or head-mounted device in a small environment. For these sequences ground-truth is available for the entire trajectory. Table \ref{table:tum_vi} shows that ORB-SLAM3 is significantly more accurate that competing approaches. The results obtained using our four sensor configurations are compared in table \ref{table:tum_vi_arvr}. The better accuracy of pure monocular compared with stereo is only apparent: the monocular solution is up-to-scale and is aligned with ground-truth with 7 DoFs, while stereo provides the true scale, and is aligned with 6 DoFs. Using monocular-inertial, we further reduce the average RMS ATE error close to 1\,cm, also obtaining the true scale. Finally, our stereo-inertial SLAM brings error below 1\,cm, making it an excellent choice for AR/VR applications.

\subsection{Multi-session SLAM}

EuRoC dataset contains several sessions for each of its three environments: 5 in Machine Hall, 3 in Vicon1 and 3 in Vicon2. To test the multi-session performance of ORB-SLAM3, we process sequentially all the sessions corresponding to each environment. Each trajectory in the same environment has ground-truth with the same world reference, which allows to perform a single global alignment to compute ATE. 

The first sequence in each room provides an initial map. Processing the following sequences starts with the creation of a new active map, that is quickly merged with the map of the previous sessions, and from that point on, ORB-SLAM3 profits from reusing the previous map.

Table \ref{table:multiple_session_experiment} reports the global multi-session RMS ATE for the four sensor configurations in the three rooms, comparing with the two only published multi-session results in EuRoC dataset: CCM-SLAM \cite{schmuck2018ccm} that reports pure monocular results in MH01-MH03, and VINS-Mono \cite{qin2018vins} in the five Machine Hall sequences, using monocular-inertial. In both cases ORB-SLAM3 more than doubles the accuracy of competing methods. In the case of VINS-Mono, ORB-SLAM3 obtains 2.6 better accuracy in single-session, and the advantage goes up to 3.2 times in multi-session, showing the superiority of our map merging operations.

Comparing these multi-session performances with the single-session results reported in Table\,\ref{table:euro_sensor_error} the most notable difference is that multi-sessions monocular and stereo SLAM can robustly process the difficult sequences V103 and V203, thanks to the exploitation of the previous map. 

\begin{figure}
\centering
  \includegraphics[width=0.95\columnwidth]{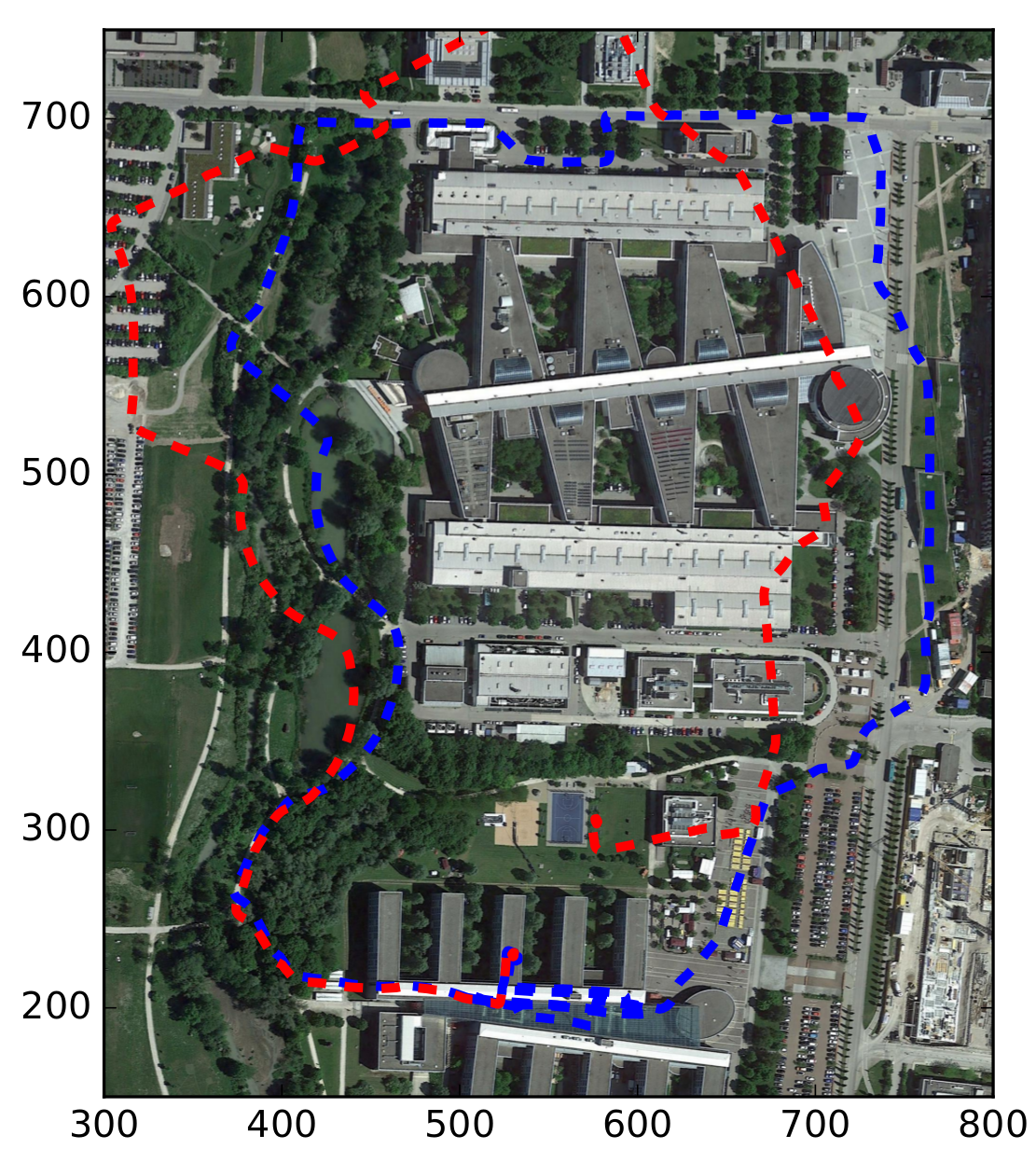}
  \caption{ Multi-session stereo-inertial. In red, the trajectory estimated after single-session processing of outdoors1. In blue, multi-session processing of magistrale2 first, and then outdoors1.}
  \label{fig:tum_outdoor}
\end{figure}

We have also performed some multi-session experiments on the TUM-VI dataset. Figure \ref{fig:tum_map} shows the result after processing several sequences inside the TUM building\footnote{Videos of this and other experiments can be found at \url{https://www.youtube.com/channel/UCXVt-kXG6T95Z4tVaYlU80Q}}. In this case, the small {\em room} sequence provides loop closures that were missing in the longer sequences, bringing all errors to centimeter level. Although ground-truth is not available outside the {\em room}, comparing the figure with the figures published in \cite{schubert2020tum} clearly shows our point: our multi-session SLAM system obtains far better accuracy that existing visual-inertial odometry systems. This is further exemplified in Figure \ref{fig:tum_outdoor}. Although ORB-SLAM3 ranks higher in stereo inertial single-session processing of outdoors1, there is still a significant drift ($\approx 60$ m). In contrast, if outdoors1 is processed after magistrale2 in a multi-session manner, this drift is significantly reduced, and the final map is much more accurate.

\subsection{Computing Time}

\begin{table*}
\centering
\caption{Running time of the main parts of our tracking and mapping threads compared to ORB-SLAM2, on EuRoC V202 (mean time and standard deviation in ms).}
\begin{tabular}{|l|l|c||c|c|c|c|} \hline
    \multirow{7}{*}{Settings} & System & ORB-SLAM2 & ORB-SLAM3 & ORB-SLAM3 & ORB-SLAM3 & ORB-SLAM3  \\ \cline{2-7}
                              & Sensor & Stereo & Monocular & Stereo & Mono-Inertial & Stereo-Inertial  \\  \cline{2-7}
                              & Resolution & 752$\times$480 & 752$\times$480 & 752$\times$480 & 752$\times$480 & 752$\times$480 \\ \cline{2-7} 
                              & Cam. FPS & 20Hz & 20Hz & 20Hz & 20Hz & 20Hz \\ \cline{2-7} 
                              & IMU & - & - & - & 200Hz & 200HZ \\ \cline{2-7}
                              & ORB Feat. & 1200 & 1000 & 1200 & 1000 & 1200 \\ \cline{2-7}
                              & RMS ATE & 0.035 & 0.029 & 0.028 & 0.021 & 0.014 \\\hline \hline
    \multirow{8}{*}{Tracking} & Stereo rect. & 3.07$\pm$0.80 & - & 1.32$\pm$0.43 & - & 1.60$\pm$0.74 \\ \cline{2-7}
         & ORB extract & 11.20$\pm$2.00 & 12.40$\pm$5.10 & 15.68$\pm$4.74 & 11.98$\pm$4.78 & 15.22$\pm$4.37 \\ \cline{2-7}
         & Stereo match & 10.38$\pm$2.57 & - & 3.35$\pm$0.92 & - & 3.38$\pm$1.07 \\ \cline{2-7}
         & IMU integr. & - & - & - & 0.18$\pm$0.11 & 0.22$\pm$0.20 \\ \cline{2-7}
         & Pose pred & 2.20$\pm$0.72 & 1.87$\pm$0.68 & 2.69$\pm$0.85 & 0.09$\pm$0.41 & 0.15$\pm$0.71 \\ \cline{2-7}
         & LM Track & 9.89$\pm$4.95 & 4.98$\pm$1.65 & 6.31$\pm$2.85 & 8.22$\pm$2.52 & 11.51$\pm$3.33 \\ \cline{2-7}
         & New KF dec & 0.20$\pm$0.43 & 0.04$\pm$0.03 & 0.12$\pm$0.19 & 0.05$\pm$0.03 & 0.18$\pm$0.25 \\ \cline{2-7}
         & Total & 37.87$\pm$7.49 & 21.52$\pm$6.45 & 31.48$\pm$5.80 & 23.22$\pm$14.98 & 33.05$\pm$9.29 \\ \hline \hline
    \multirow{6}{*}{Mapping} & KF Insert & 8.72$\pm$3.60 & 9.25$\pm$4.62 & 8.03$\pm$2.96 & 13.17$\pm$7.43 & 8.53$\pm$2.17 \\ \cline{2-7}
         & MP Culling & 0.25$\pm$0.09 & 0.09$\pm$0.04 & 0.32$\pm$0.15 & 0.07$\pm$0.04 & 0.24$\pm$0.24 \\ \cline{2-7}
         & MP Creation & 36.88$\pm$14.53 & 22.78$\pm$8.80 & 18.23$\pm$9.84 & 30.19$\pm$12.95 & 23.88$\pm$9.97 \\ \cline{2-7}
         & LBA & 139.61$\pm$124.92 & 216.95$\pm$188.77 & 134.60$\pm$136.28 & 121.09$\pm$44.81 & 152.70$\pm$38.37 \\ \cline{2-7}
         & KF Culling & 4.37$\pm$4.73 & 18.88$\pm$12.217 & 5.49$\pm$5.09 & 26.25$\pm$17.08 & 11.15$\pm$7.67 \\ \cline{2-7}
         & Total & 173.81$\pm$139.07 & 266.61$\pm$207.80 & 158.84$\pm$147.84 & 191.50$\pm$80.54 & 196.61$\pm$54.52 \\ \hline \hline
    \multirow{2}{*}{Map Size} & KFs & 278 & 272 & 259 & 332 & 135 \\ \cline{2-7}
        & MPs & 14593 & 9686 & 14245 & 10306 & 9761 \\ 
        \hline 
    \end{tabular}
\label{table:timing1}
\end{table*}

\begin{table*}
\centering
\caption{Running time of the main operations for loop closing and map merging for a multisesion experiment on sequences V201, V202 and V203 from EuRoC dataset (mean time and standard deviation in ms).} 
\begin{tabular}{|l|l|c|c|c|c|} \hline
    \multirow{6}{*}{Settings} & Sensor & Monocular & Stereo & Mono-Inertial & Stereo-Inertial  \\  \cline{2-6}
                              & Resolution & 752$\times$480 & 752$\times$480 & 752$\times$480 & 752$\times$480 \\ \cline{2-6} 
                              & Cam. FPS & 20Hz & 20Hz & 20Hz & 20Hz \\ \cline{2-6} 
                              & IMU & - & - & 200Hz & 200HZ \\ \cline{2-6}
                              & ORB Feat. & 1000 & 1200 & 1000 & 1200 \\ \cline{2-6}
                              & RMS ATE & 0.284 & 0.163 & 0.048 & 0.046 \\ \hline \hline
    \multirow{3}{*}{Place Recognition} & Database query & 0.96$\pm$0.58 & 1.06$\pm$0.58 & 1.04$\pm$0.59 & 1.02$\pm$0.60 \\ \cline{2-6}
         & Compute Sim3/SE3 & 3.61$\pm$2.81 & 5.26$\pm$3.79 & 2.98$\pm$2.26 & 5.71$\pm$3.54 \\ \cline{2-6}
         & Total & 3.92$\pm$3.28 & 5.26$\pm$4.39 & 3.45$\pm$2.81 & 5.89$\pm$4.29 \\  \hline \hline
    \multirow{4}{*}{Map Merging} & Merge Maps & 152.03$\pm$45.85 & 68.56$\pm$13.56 & 129.08$\pm$8.26 & 91.07$\pm$5.56 \\ \cline{2-6}
    & Welding BA & 52.09$\pm$14.08 & 35.57$\pm$7.94 & 103.14$\pm$6.08 & 58.15$\pm$4.84 \\ \cline{2-6}
    & Opt. Essential Graph & 5.82$\pm$3.01 & 10.98$\pm$9.79 & 52.83$\pm$17.81 & 36.08$\pm$17.95 \\ \cline{2-6}
    & Total & 221.90$\pm$58.73 & 120.63$\pm$16.23 & 287.33$\pm$15.58 & 187.82$\pm$6.38 \\ \hline
    \multirow{3}{*}{Merge info}& \# Detected merges & 5 & 4 & 2 & 2 \\ \cline{2-6}
    & Merge size (\# keyframes) & 31$\pm$1 & 31$\pm$3 & 25$\pm$1 & 25$\pm$0 \\ \cline{2-6}
    & Merge size (\# map points) & 2476$\pm$207 & 2697$\pm$718 & 2425$\pm$88 & 4260$\pm$160 \\ \hline \hline
    \multirow{3}{*}{Loop} & Loop Fusion & 311.82$\pm$333.49 & 29.07$\pm$23.64 & - & 25.67 \\ \cline{2-6}
    & Opt. Essential Graph & 254.84$\pm$87.03 & 84.36$\pm$37.56 & - & 95.13 \\ \cline{2-6}
    & Total & 570.39$\pm$420.77 & 118.62$\pm$59.93 & - & 124.77 \\ \hline
    \multirow{2}{*}{Loop info}& \# Detected loops & 3 & 4 & 0 & 1 \\ \cline{2-6}
    & Loop size (\# keyframes) & 58$\pm$60 & 27$\pm$9 & - & 60 \\ \hline
    \hline
    \multirow{5}{*}{Loop Full BA} & Full BA & 4010.14$\pm$1835.85 & 1118.54$\pm$563.75 & - & 1366.64 \\ \cline{2-6}
         & Map Update & 124.80$\pm$6.07 & 13.65$\pm$12.86 & - & 163.06 \\ \cline{2-6}
         & Total & 4134.94$\pm$1829.78 & 1132.19$\pm$572.28 & - & 1529.69 \\ \cline{2-6} \cline{2-6}
         & BA size (\# keyframes) & 345$\pm$147 & 220$\pm$110 & - & 151 \\ \cline{2-6}
         & BA size (\# map points) & 13511$\pm$3778 & 12297$\pm$4572 & - & 14397 \\\hline 
    \end{tabular}
\label{table:timing2}
\end{table*}

Table \ref{table:timing1} summarizes the running time of the main operations performed in the tracking and mapping threads, showing that our system is able to run in real time at 30-40 frames and at 3-6 keyframes per second. The inertial part takes negligible time during tracking and, in fact can render the system more efficient as the frame rate could be safely reduced. In the mapping thread, the higher number of variables per keyframe has been compensated with a smaller number of keyframes in the inertial local BA, achieving better accuracy, with similar running time. As the tracking and mapping threads work always in the active map, multi-mapping  does not introduce significant overhead.

Table \ref{table:timing2} summarizes the running time of the main steps for loop closing and map merging. The novel place recognition method only takes 10\,ms per keyframe. Times for merging and loop closing remain below one second, running only a pose-graph optimization. For loop closing, performing a full bundle adjustment may increase times up to a few seconds, depending on the size of the involved maps. In any case, as both operations are executed in a separate thread (Fig. \ref{fig:orbslam3_workflow}) they do not interfere with the real time performance of the rest of the system. The visual-inertial systems perform just two map merges to join three sequences, while visual systems perform some additional merges to recover from tracking losses. Thanks to their lower drift, visual-inertial systems also perform less loop closing operations compared with pure visual systems.

Although it would be interesting, we do not compare running time against other systems, since this would require a significant effort that is beyond the scope of this work.


\section{Conclusions}
Building on \cite{mur2015orb,mur2017orb,mur2017visual}, we have presented ORB-SLAM3, the most complete open-source library for visual, visual-inertial and multi-session SLAM, with monocular, stereo, RGB-D, pin-hole and fisheye cameras. Our main contributions, apart from the integrated library itself, are the fast and accurate IMU initialization technique, and the multi-session map-merging functions, that rely on an new place recognition technique with improved recall.

Our experimental results show that ORB-SLAM3 is the first visual and visual-inertial system capable of effectively exploiting short-term, mid-term, long-term and multi-map data associations, reaching an accuracy level that is beyond the reach of existing systems. Our results also suggest that, regarding accuracy, the capability of using all these types of data association overpowers other choices such as using direct methods instead of features, or performing keyframe marginalization for local BA, instead of assuming an outer set of static keyframes as we do. 

The main failure case of ORB-SLAM3 is low-texture environments. Direct methods are more robust to low-texture, but are limited to short-term \cite{Engel-et-al-pami2018} and mid-term \cite{zubizarreta2020direct} data association. On the other hand, matching feature descriptors successfully solves long-term and multi-map data association, but seems to be less robust for tracking than Lucas-Kanade, that uses photometric information. An interesting line of research could be developing photometric techniques adequate for the four data association problems. We are currently exploring this idea for map building from endoscope images inside the human body.

About the four different sensor configurations, there is no question, stereo-inertial SLAM provides the most robust and accurate solution. Furthermore, the inertial sensor allows to estimate pose at IMU rate, which is orders of magnitude higher than frame rate, being a key feature for some use cases. For applications where a stereo camera is undesirable because of its higher bulk, cost, or processing requirements, you can use monocular-inertial without missing much in terms of robustness and accuracy. Only keep in mind that pure rotations during exploration would not allow to estimate depth. 

In applications with slow motions, or without roll and pitch rotations, such as a car in a flat area, IMU sensors can be difficult to initialize. In those cases, if possible, use stereo SLAM. Otherwise, recent advances on depth estimation from a single image with CNNs offer good promise for reliable and true-scale monocular SLAM \cite{yang2020d3vo}, at least in the same type of environments where the CNN has been trained.

\bibliographystyle{IEEEtran}
\bibliography{./references,mingo,IEEEabrv}

\begin{IEEEbiography}[{\includegraphics[width=1in,height=1.25in,clip,keepaspectratio]{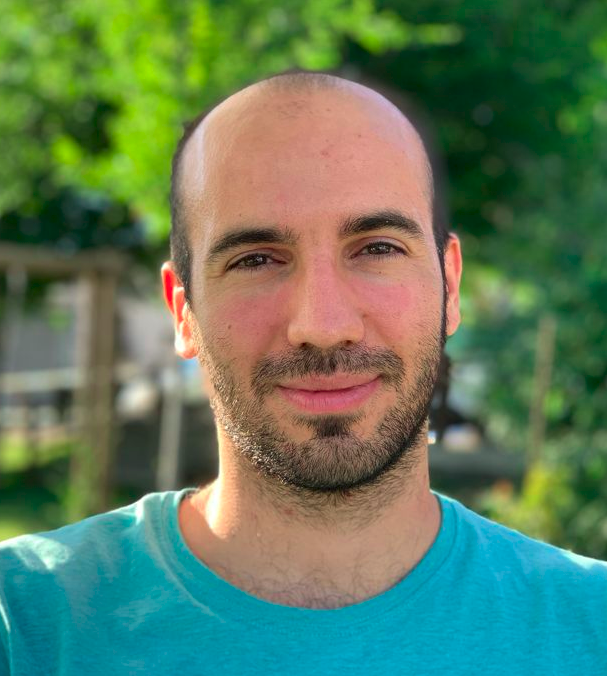}}]%
{Carlos Campos} received an Electronic Engineering degree (mention in Signal Processing) from INP-Toulouse and the Industrial Engineering Bachelor and M.S. degree (mention in Robotics and Computer Vision) from the University of Zaragoza. He is currently working towards the PhD. degree with the I3A Robotics, Perception and Real-Time Group. His research interests include Visual-Inertial Localization and Mapping for AR/VR applications. 
\end{IEEEbiography}

\begin{IEEEbiography}[{\includegraphics[width=1in,height=1.25in,clip,keepaspectratio]{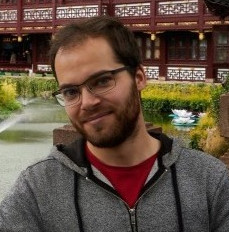}}]%
{Richard Elvira} received a Bachelor's Degree in Informatics Engineering (mention in Computing) and Master's in Biomedical Engineering (mention in Information and Communication Technologies in Biomedical Engineering) from Universidad de Zaragoza, where he is currently PhD. student in the I3A Robotics, Perception and Real-Time Group. His research interests are real-time visual SLAM and place recognition in rigid environments.
\end{IEEEbiography}

\begin{IEEEbiography}[{\includegraphics[width=1in,height=1.25in,clip,keepaspectratio]{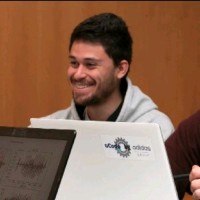}}]%
{Juan J. Gómez Rodríguez} received a Bachelor's Degree in Informatics Engineering (mention in Computing) and Master's in Biomedical Engineering (mention in Information and Communication Technologies in Biomedical Engineering) from Universidad de Zaragoza, where he is currently working towards the PhD. degree with the I3A Robotics, Perception and Real-Time Group. His research interests are real-time visual SLAM for both rigid and deformable environments.
\end{IEEEbiography}

\begin{IEEEbiography}[{\includegraphics[width=1in,height=1.25in,clip,keepaspectratio]{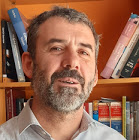}}]%
{J. M. Martínez Montiel} (Arnedo, Spain, 1967) received the M.S. and PhD degrees in electrical engineering from Universidad de Zaragoza, Spain, in 1992 and 1996, respectively. He has been awarded several Spanish MEC grants to fund research with the University of Oxford, U.K., and Imperial College London, U.K.

He is currently a full professor with the Departamento de Informática e Ingeniería de Sistemas, Universidad de Zaragoza, where he is in charge of perception and computer vision research grants and courses. His interests include real-time visual SLAM for rigid and non-rigid environments, and the transference of this technology to robotic and non-robotic application domains. He has received several awards, including the 2015 King-Sun Fu Memorial IEEE Transactions on Robotics Best Paper Award. Since 2020 he coordinates the EU FET EndoMapper grant to bring visual SLAM to intracorporeal medical scenes.
\end{IEEEbiography}

\begin{IEEEbiography}[{\includegraphics[width=1in,height=1.25in,clip,keepaspectratio]{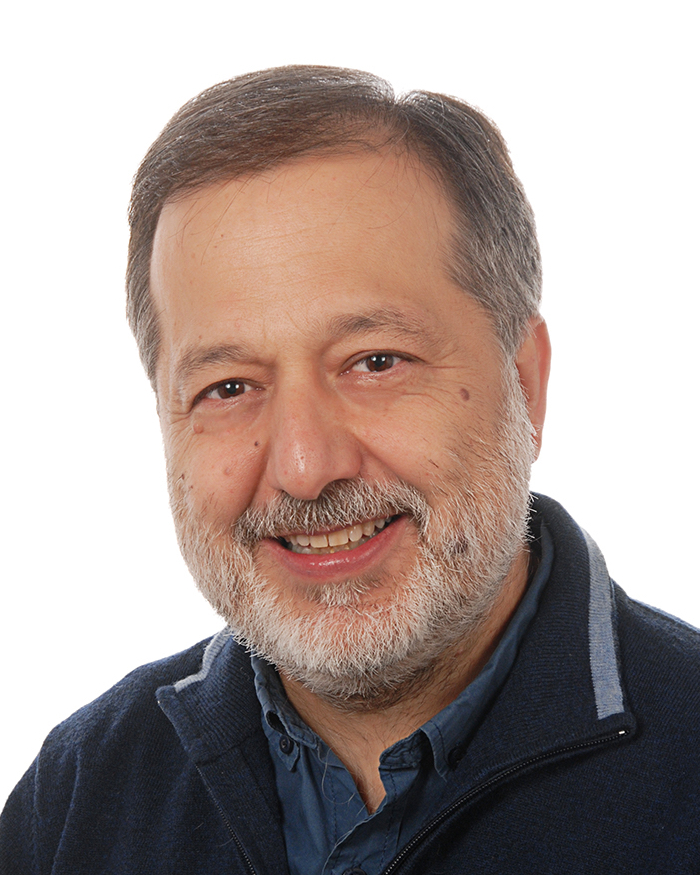}}]{Juan D. Tard\'os}
(Huesca, Spain, 1961) received the M.S. and Ph.D. degrees in electrical
engineering from the University of Zaragoza, Spain, in 1985 and
1991, respectively. He is Full Professor with the Departamento de
Inform\'atica e Ingenier\'ia de Sistemas, University of Zaragoza,
where he is in charge of courses in robotics, computer vision, and
artificial intelligence. His research interests include
SLAM, perception and mobile robotics. He received the 2015 King-Sun Fu Memorial IEEE Transactions on Robotics Best Paper Award, for the paper describing the monocular SLAM system ORB-SLAM.
\end{IEEEbiography}

\end{document}